\documentclass[sn-basic]{sn-jnl}
\usepackage{graphicx}%
\usepackage{multirow}%
\usepackage{amsmath,amssymb,amsfonts}%
\usepackage{amsthm}%
\usepackage{mathrsfs}%
\usepackage[title]{appendix}%
\usepackage{xcolor}%
\usepackage{textcomp}%
\usepackage{manyfoot}%
\usepackage{booktabs}%
\usepackage{algorithm}%
\usepackage{algorithmicx}%
\usepackage{algpseudocode}%
\usepackage{listings}%
\usepackage{gensymb}%
\usepackage{array}%
\usepackage{natbib}%
\usepackage{graphicx} % for resizing table
\usepackage{multirow}
\usepackage{multicol}
\usepackage{longtable}
\usepackage{tabularx}
\usepackage{mathtools}
\usepackage{amsmath}
\usepackage{amssymb}  % For Blackboard fonts, used in Set of natural numbers
\usepackage{amsfonts}
\usepackage{gensymb} %For degree sysmbol
\usepackage{booktabs} % for better looking table
\usepackage{threeparttable}  % for table footnotes
\usepackage{caption}    % for better caption formatting
\usepackage{rotating} % For sidewaystable
\usepackage{tabularx}
\usepackage{booktabs}
\usepackage{makecell}
\theoremstyle{thmstyleone}%
\theoremstyle{thmstyletwo}%

\theoremstyle{thmstylethree}%

\begin{document}

\title[Article Title]{PollutionNet: A Vision Transformer Framework for Climatological Assessment of NO\textsubscript{2} and SO\textsubscript{2} Using Satellite-Ground Data Fusion}


\author*[1,3]{\fnm{Prasanjit} \sur{Dey}}\email{d22124678@mytudublin.ie}

\author[2,3]{\fnm{Soumyabrata} \sur{Dev}}\email{soumyabrata.dev@ucd.ie}

\author[1,3]{\fnm{Bianca} \sur{Schoen-Phelan}}\email{bianca.schoenphelan@tudublin.ie}

\affil*[1]{\orgdiv{School of Computer Science}, \orgname{Technological University Dublin}, \orgaddress{\country{Ireland}}}

\affil[2]{\orgdiv{School of Computer Science}, \orgname{University College Dublin}, \orgaddress{\country{Ireland}}}

\affil[3]{\orgname{ADAPT Research Ireland Centre}, \orgaddress{\country{Ireland}}}

\abstract{

Accurate assessment of atmospheric nitrogen dioxide (NO\textsubscript{2}) and sulfur dioxide (SO\textsubscript{2}) is essential for understanding climate-air quality interactions, supporting environmental policy, and protecting public health. Traditional monitoring approaches face limitations: satellite observations provide broad spatial coverage but suffer from data gaps, while ground-based sensors offer high temporal resolution but limited spatial extent. To address these challenges, we propose PollutionNet, a Vision Transformer-based framework that integrates Sentinel-5P TROPOMI vertical column density (VCD) data with ground-level observations. By leveraging self-attention mechanisms, PollutionNet captures complex spatiotemporal dependencies that are often missed by conventional CNN and RNN models. Applied to Ireland (2020-2021), our case study demonstrates that PollutionNet achieves state-of-the-art performance (RMSE: 6.89~\textmu g/m\textsuperscript{3} for NO\textsubscript{2}, 4.49~\textmu g/m\textsuperscript{3} for SO\textsubscript{2}), reducing prediction errors by up to 14\% compared to baseline models. Beyond accuracy gains, PollutionNet provides a scalable and data-efficient tool for applied climatology, enabling robust pollution assessments in regions with sparse monitoring networks. These results highlight the potential of advanced machine learning approaches to enhance climate-related air quality research, inform environmental management, and support sustainable policy decisions. The code and data used in this study are publicly available at: \url{https://github.com/Prasanjit-Dey/PollutionNet}.
}

\keywords{Applied climatology, Atmospheric pollution, Air quality monitoring, Satellite observation, Vision Transformer (ViT)}



\maketitle
\section{Introduction}
\label{sec:introduction}

Atmospheric nitrogen dioxide (NO\textsubscript{2}) and sulfur dioxide (SO\textsubscript{2}) are key pollutants emitted from industrial activities, transportation, and energy production, contributing to smog formation, acid rain, and adverse health effects~\cite{shikwambana2020trend, gao2023assessing}. Monitoring these gases is critical for environmental and public health policymaking, yet their dynamic spatiotemporal variability poses significant challenges for accurate assessment~\cite{rafaj2018outlook, tamehri2023impact}.

Current monitoring relies on two primary data sources: (1) ground-based stations, which provide high temporal resolution but lack spatial coverage, especially in remote regions~\cite{wu2022boosting}, and (2) satellite observations (e.g., TROPOMI/Sentinel-5P), which offer global coverage but suffer from data gaps due to cloud cover, nighttime limitations, and retrieval artifacts~\cite{li2020version, kazemi2023monitoring}. While machine learning models like CNNs and RNNs have been applied to fuse these data sources, their ability to capture long-range dependencies and complex spatial patterns remains limited. CNNs excel at local feature extraction but struggle with global context, while RNNs face computational inefficiencies in modeling long-term trends~\cite{zhang2022deep, dua2019real}. Hybrid architectures attempt to bridge this gap but often introduce complexity without commensurate gains in performance.

Vision Transformers (ViTs) present a promising alternative, leveraging self-attention mechanisms to model global relationships in data without the inductive biases of CNNs or RNNs. Their ability to process multi-scale features and handle missing data makes them particularly suited for integrating heterogeneous inputs like satellite vertical column density (VCD) maps and ground sensor readings. However, their potential for atmospheric pollution assessment remains underexplored, with most studies still relying on conventional deep learning approaches.

In this case study, we propose PollutionNet, a ViT-based framework designed to assess NO\textsubscript{2} and SO\textsubscript{2} pollution by synergistically combining TROPOMI satellite data and ground-level observations. Our work addresses three key gaps: (1) the lack of methods leveraging ViTs for trace gas prediction, (2) the need for robust handling of satellite data gaps, and (3) the integration of multi-source data to improve spatial generalizability.

\subsection*{Contributions}

\begin{itemize}
    \item \textbf{ViT for pollution assessment:} We introduce PollutionNet, the first Vision Transformer-based model tailored to predict surface-level NO\textsubscript{2} and SO\textsubscript{2} concentrations using both satellite and ground-based data, demonstrating superior performance over CNNs/RNNs.
    \item \textbf{Case study validation:} Through a comprehensive evaluation using TROPOMI VCDs and ground observations, we show PollutionNet achieves state-of-the-art results (RMSE: 6.89~$\mu$g/m\textsuperscript{3} for NO\textsubscript{2}, 4.49~$\mu$g/m\textsuperscript{3} for SO\textsubscript{2}), addressing real-world data gaps.
    \item \textbf{Reproducibility:} We release all code and processing pipelines to facilitate future research in air quality modeling.
\end{itemize}

The paper is structured as follows: Section~\ref{sec:related} reviews prior work; Section~\ref{sec:study} details the study area and data; Section~\ref{sec:method} presents PollutionNet’s architecture; Section~\ref{sec:result} discusses results; and Section~\ref{conclusion} outlines future directions.
%------------------------------------------------------%

\section{Related Works} \label{sec:related}

Recent advances in air pollution modeling leverage either ground-based or satellite-based data, each with distinct trade-offs in spatiotemporal coverage and resolution. We categorize existing approaches into three groups: (1) ground observation methods, (2) satellite-driven models, and (3) emerging ViT applications in environmental science.

\subsection{Ground-Based Approaches}

Accurate assessment of atmospheric NO\textsubscript{2} and SO\textsubscript{2} pollution has been approached through two primary data sources: ground-based monitoring and satellite observations. Ground-based methods rely on sensor networks that provide high temporal resolution but suffer from limited spatial coverage, restricting their use to urban or well-monitored regions. Early studies employed machine learning techniques such as random forests and support vector machines (SVMs) to predict pollutant concentrations, achieving moderate accuracy (RMSE: 10-12~$\mu$g/m\textsuperscript{3}) but struggling with generalizability beyond local areas~\cite{masih2019application, shaban2016urban}. More recent advances introduced recurrent architectures like LSTMs and Bi-GRUs to better model temporal dependencies, though these methods still faced challenges in capturing long-term trends and cross-regional patterns~\cite{hamami2020univariate, dairi2021integrated}. Hybrid CNN-LSTM models attempted to combine spatial and temporal learning but were computationally intensive and often limited to specific urban environments~\cite{zhang2022deep}.

\subsection{Satellite-Based Approaches}

Satellite-based approaches, such as those using TROPOMI/Sentinel-5P data, offer broader spatial coverage but contend with data gaps due to cloud cover, nighttime limitations, and retrieval artifacts. Tree-based models (e.g., LightGBM) trained on satellite-derived VCDs achieved competitive results (RMSE: $\sim$8.5~$\mu$g/m\textsuperscript{3} for NO\textsubscript{2} in China) but often lacked integration with ground-level validation~\cite{long2022estimating, wang2021estimating}. Deep neural networks (DNNs) were also applied to fuse satellite and meteorological data, yet their reliance on conventional architectures limited their ability to capture long-range spatial dependencies~\cite{li2021spatiotemporal, chan2021estimation}. Multi-model ensembles further improved robustness but introduced complexity in calibration and deployment~\cite{rowley2023predicting}.

\subsection{Vision Transformers in Environmental Science}
ViTs have emerged as a powerful alternative in environmental science due to their ability to model global relationships through self-attention mechanisms. In land cover classification, ViTs outperformed CNNs by capturing large-scale spatial patterns in satellite imagery~\cite{yao2023extended}. Similarly, climate modeling studies demonstrated ViTs’ effectiveness in processing high-resolution climate data for temperature and precipitation forecasting~\cite{lin2023mmst, nguyen2024climatelearn}. However, their application to air quality prediction—particularly for NO\textsubscript{2} and SO\textsubscript{2} remains underexplored. They hold strong potential to integrate multi-source data and resolve complex spatiotemporal interactions.

\section{Study Area and Dataset Preparation}
\label{sec:study}

\begin{figure}[!ht]
\centering
\includegraphics[width=0.70\linewidth]{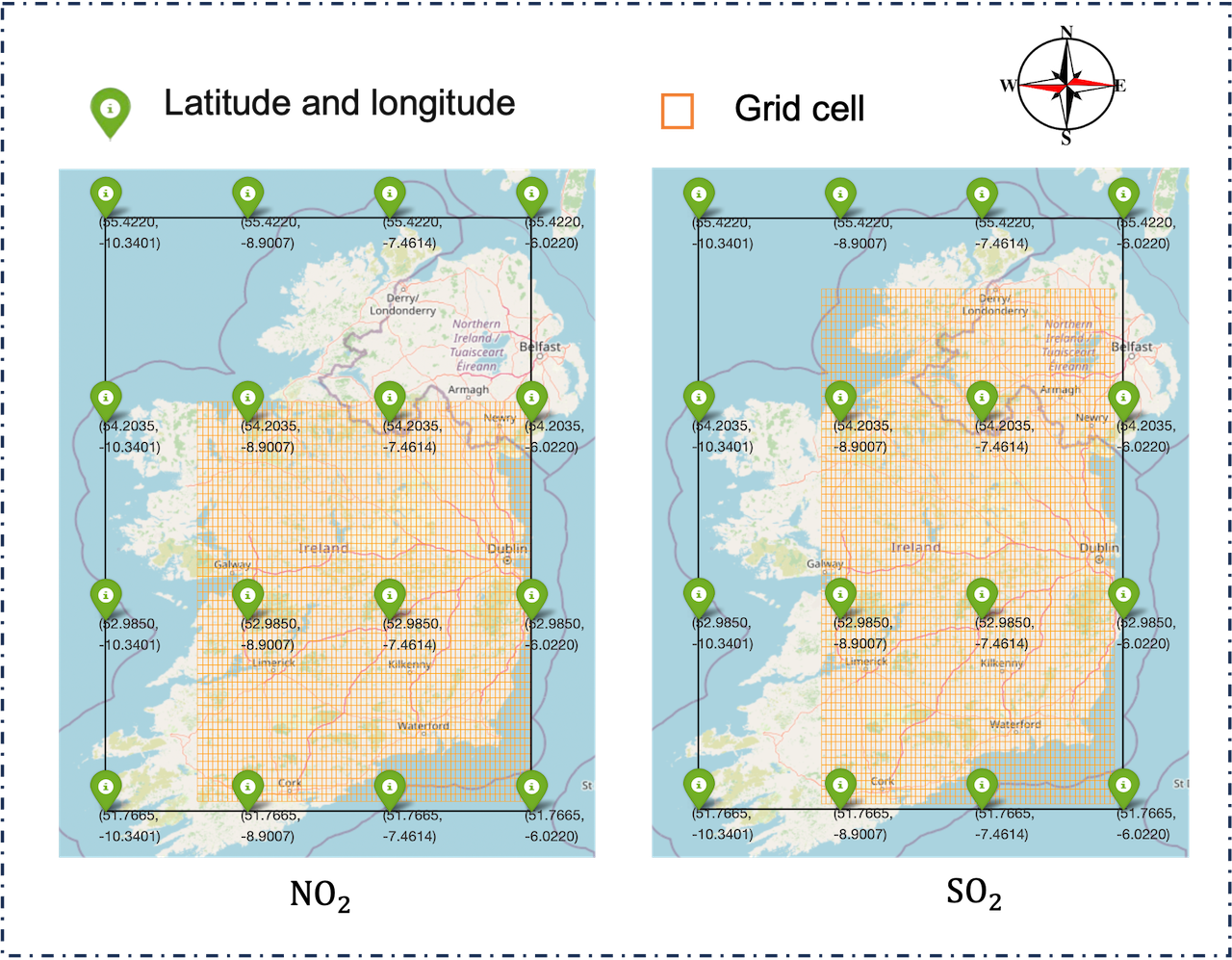}
\caption{Study region for satellite and ground observations of NO\textsubscript{2} and SO\textsubscript{2} concentrations. Grid cells represent the spatial domains for each pollutant.}
\label{fig:area}
\end{figure}

This study examines near-surface concentrations of nitrogen dioxide (NO\textsubscript{2}) and sulfur dioxide (SO\textsubscript{2}) over Ireland using ground-based and satellite-derived datasets from January 1, 2020, to May 1, 2021. The study area, illustrated in Fig.~\ref{fig:area}, was defined using distinct geographical boundaries for each pollutant to account for their differing emission sources and monitoring station distributions.

\subsection{Geographical and Grid Configuration}

\subsubsection*{Spatial Domains}
The study area for NO\textsubscript{2} was bounded by its southwestern (51.795$\degree$N, $-9.089\degree$E) and northeastern (54.323$\degree$N, $-6.032\degree$E) edges, encompassing urban regions with high traffic and industrial activity, including Dublin, Cork, and Limerick. These areas were selected due to their dense network of ground monitoring stations and significant NO\textsubscript{2} emission sources.

For SO\textsubscript{2}, the domain was defined by its southwestern (51.795$\degree$N, $-9.089\degree$E) and northeastern (55.004$\degree$N, $-6.105\degree$E) corners, covering industrial zones such as power plants and manufacturing facilities, where SO\textsubscript{2} emissions are most prevalent.

\subsubsection*{Grid Design}
Both pollutants were analyzed using a 0.05$\degree \times 0.05\degree$ spatial resolution grid to balance computational efficiency with sufficient spatial detail. However, the grid dimensions differed to align with the distinct spatial distributions of each pollutant.

The NO\textsubscript{2} grid consisted of 49 rows $\times$ 67 columns, optimized to capture fine-scale variations in urban areas. In contrast, the SO\textsubscript{2} grid was structured as 64 rows $\times$ 59 columns, reflecting the broader spatial extent of industrial emissions. This approach ensured that the analysis accurately represented the unique dispersion patterns of each pollutant.

\subsection{Satellite-Observed NO\textsubscript{2} and SO\textsubscript{2} Data}

We obtained NO\textsubscript{2} and SO\textsubscript{2} VCD measurements from the TROPOMI instrument aboard Sentinel-5P, which provides high-resolution atmospheric composition data. The satellite employs a nadir-viewing push-broom configuration, covering a 2600 km swath with spectral measurements from ultraviolet to shortwave infrared.

The original TROPOMI data, available at varying resolutions, were uniformly regridded to 0.05$\degree \times 0.05\degree$ to match our study’s requirements. The VCD values were derived using differential optical absorption spectroscopy (DOAS) algorithms and stored in netCDF format. Daily mean concentrations were extracted and restructured into geospatial matrices, resulting in 485 temporal instances for each pollutant, with dimensions matching their respective study grids.

\subsection{Ground-Observed NO\textsubscript{2} and SO\textsubscript{2} Data}

Ground-level concentration data were collected from 29 NO\textsubscript{2} monitoring stations and 14 SO\textsubscript{2} stations, operated by Ireland’s environmental regulatory authority. These measurements spanned the same period as the satellite observations (January 2020 – May 2021).

To ensure consistency with the satellite data, ground observations were spatially regridded using a nearest-neighbor interpolation approach. Each monitoring station’s measurements were assigned to the closest grid cell within the predefined 0.05$\degree \times 0.05\degree$ resolution domain. This process generated 485 daily-averaged concentration matrices for each pollutant, with dimensions of 49 $\times$ 67 for NO\textsubscript{2} and 64 $\times$ 59 for SO\textsubscript{2}, aligning with their respective satellite-derived datasets.

%%%%%%%%%%%%%%%%%%%%%%%%%%%%%%%%%%%%%%%%%%%%%%%%%%%%%%%%%%%%%%%%%%%%%%%%%%

\section{Proposed Method} \label{sec:method}
\begin{figure}[!ht]
    \centering
    \includegraphics[width=.95\textwidth]{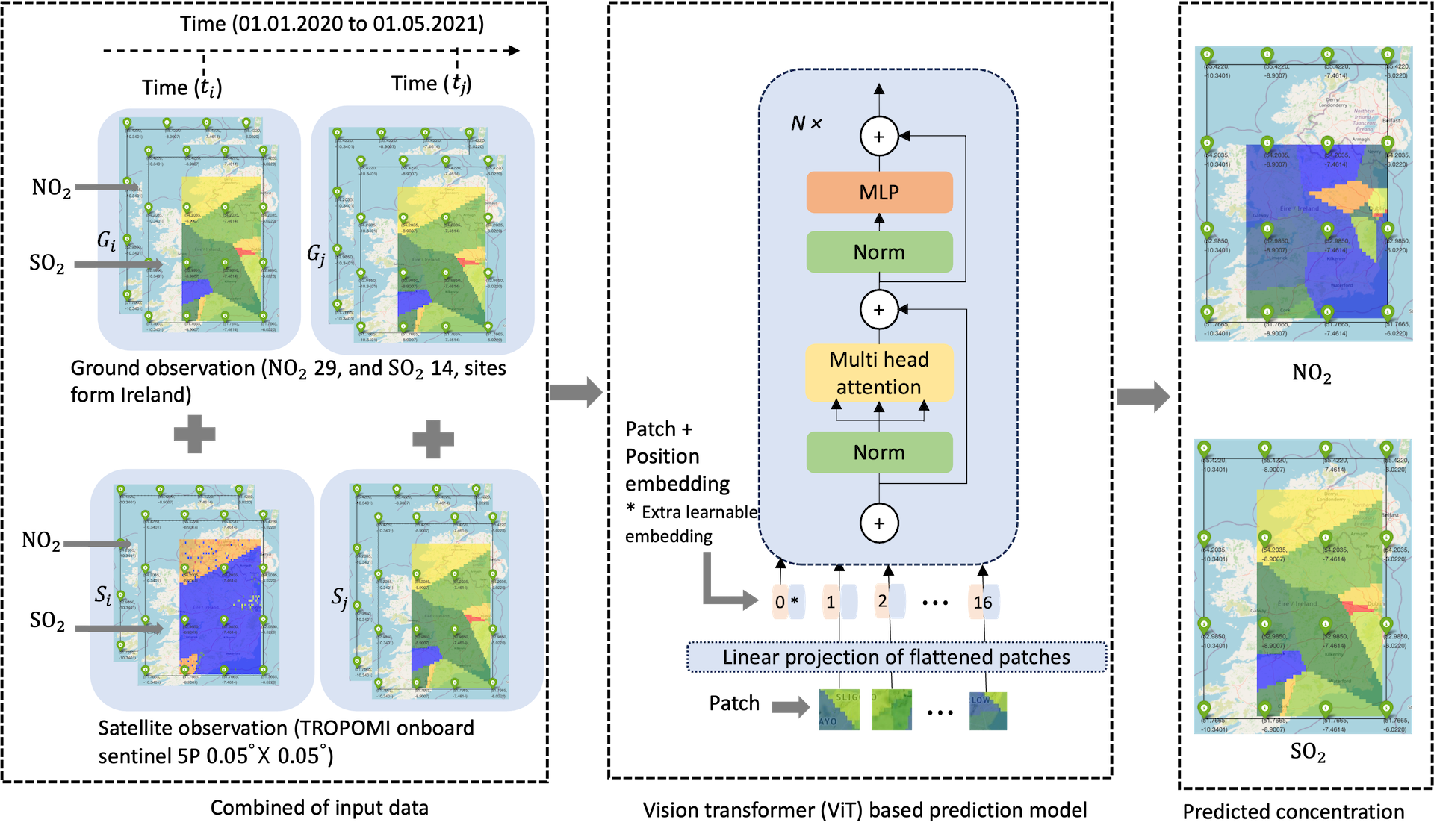}
    \caption{The Proposed frameworks of the PollutionNet for estimating and prediction of NO$_2$, and SO$_2$ concentration.}
    \label{fig:framework}
\end{figure}

This study presents a two-stage framework for forecasting near-surface NO$_2$ and SO$_2$ concentrations (Fig.~\ref{fig:framework}). First, we perform spatial-temporal fusion between satellite and ground observations to address data gaps. Second, we employ a Vision Transformer (ViT) model for concentration prediction. The entire process uses five-fold cross-validation to ensure robust model evaluation.

\subsection{Spatial-Temporal Fusion for Gap-Filling}

\begin{figure}[h!]
\centering
\includegraphics[width=0.70\textwidth]{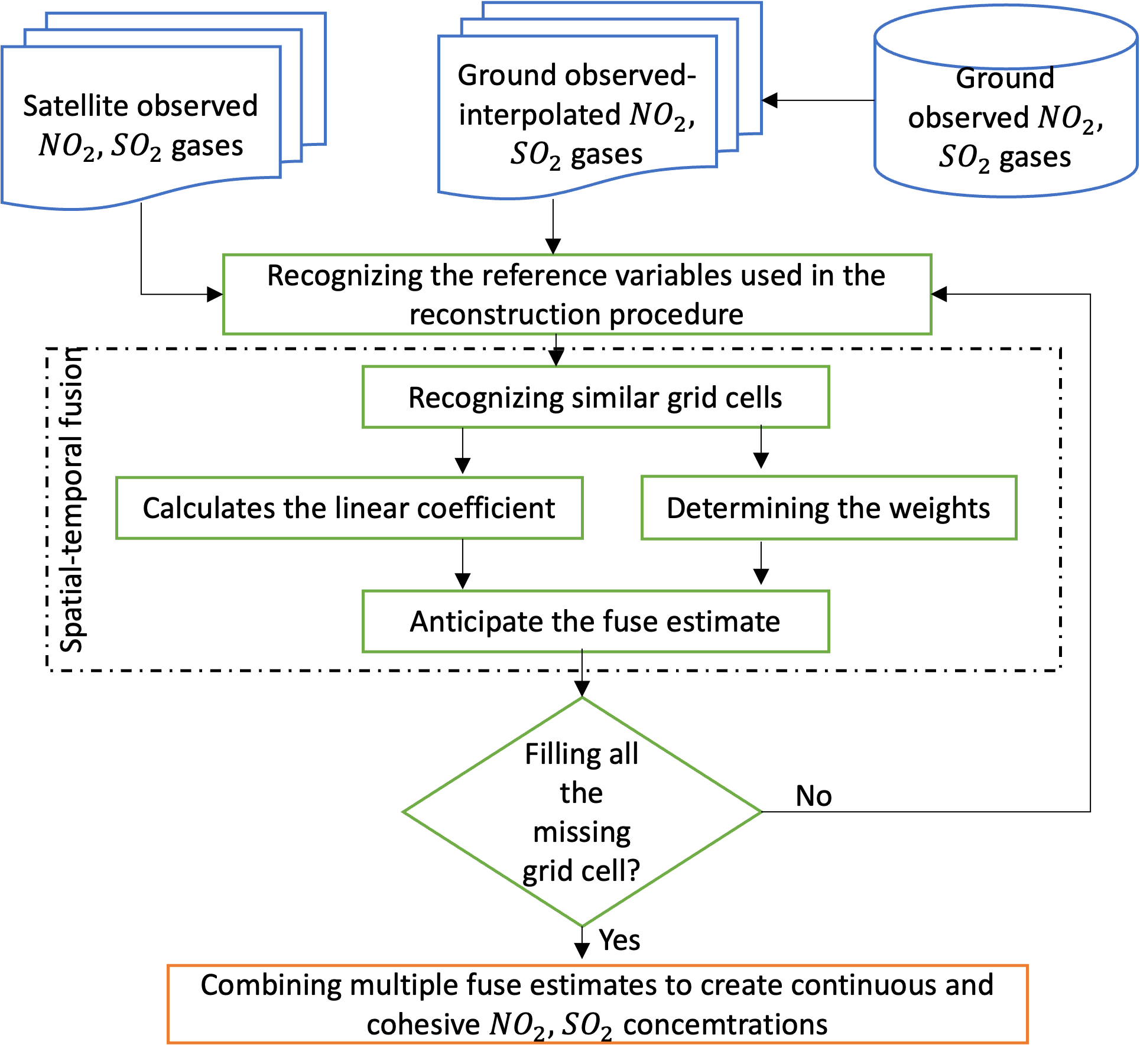}
\caption{End-to-end spatial-temporal fusion workflow showing: (1) input data acquisition from TROPOMI satellite and ground stations, (2) quality control and preprocessing, (3) core fusion algorithm execution, and (4) gap-filled output generation.}
\label{fig:flowFlow}
\end{figure}

Spatial-temporal fusion addresses the critical challenge of missing data in atmospheric monitoring by integrating complementary satellite and ground-based observations. As illustrated in Fig.~\ref{fig:flowFlow}, this methodology systematically combines satellite-derived vertical column VCDs with ground measurements to reconstruct complete datasets for NO\textsubscript{2} and SO\textsubscript{2} prediction. The fusion process overcomes limitations inherent to each data source: satellite observations provide high spatial resolution but suffer from temporal gaps due to nighttime unavailability and cloud cover, while ground stations offer continuous temporal coverage but are spatially limited to monitoring locations.

\subsubsection{Fusion Framework}

\begin{figure}[h!]
\centering
\includegraphics[width=0.50\textwidth]{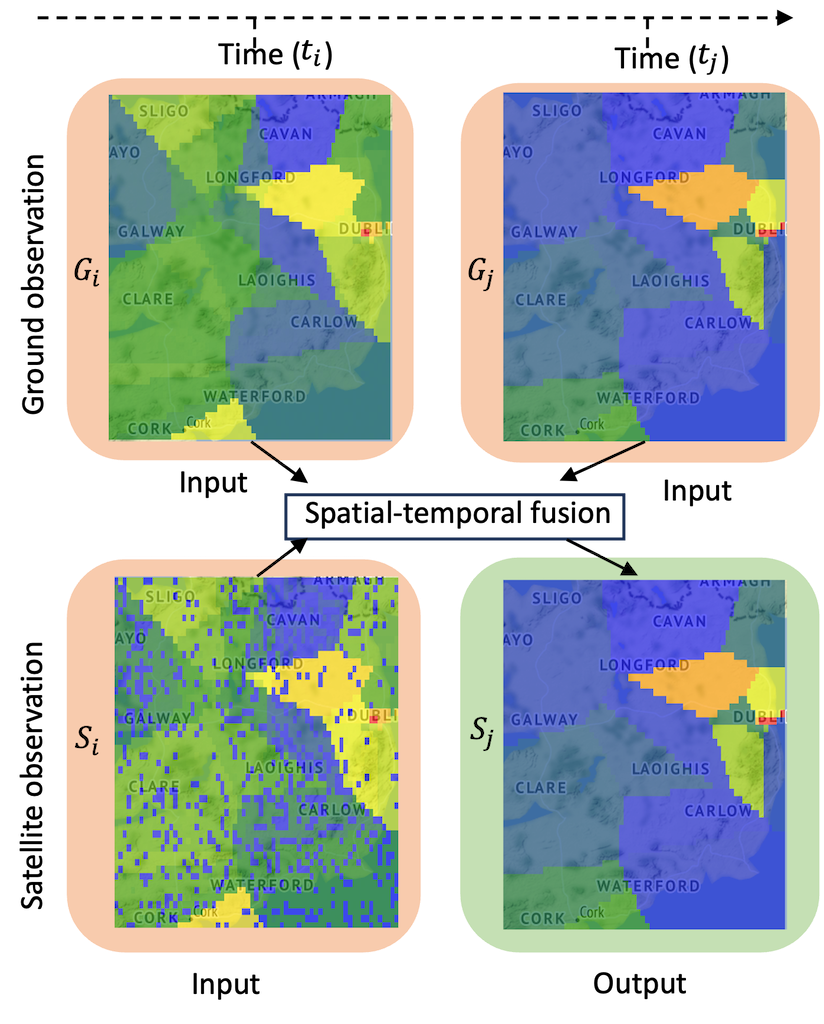}
\caption{Technical implementation of the fusion algorithm showing: (a) neighborhood selection criteria based on spatial similarity thresholds, (b) weight allocation methodology, and (c) reconstruction results for complete and partial gap scenarios.}
\label{fig:fusion}
\end{figure}

The fusion framework employs an optimized inverse distance weighting (IDW) model~\cite{li2017estimating} to combine the strengths of both data sources. Satellite data (denoted as $S$) capture fine-scale spatial patterns of pollutant distribution but contain temporal discontinuities. Conversely, ground observations ($G$) provide continuous measurements at fixed locations but lack spatial granularity. This complementary relationship is visually demonstrated in Fig.~\ref{fig:fusion}, where the fusion process bridges spatial and temporal gaps through a three-stage approach.

\subsubsection{Mathematical Formulation}

The fusion algorithm operates through three hierarchical stages. First, linear temporal projection estimates missing concentrations $S_j(x,y)$ at time $t_j$ using valid satellite observations $S_i(x,y)$ from a prior time $t_i$:

\begin{equation}
S_j(x,y) = a(x,y) \cdot S_i(x,y) + b(x,y)
\end{equation}

where coefficients $a(x,y)$ and $b(x,y)$ capture local temporal dynamics between $t_i$ and $t_j$. To enhance robustness, spatial neighborhood enhancement incorporates information from similar grid cells $(x_k, y_k)$:

\begin{equation}
S_j^i(x,y) = \sum_{k=1}^{N} w(x_k,y_k) \left[ a(x_k,y_k) \cdot S_i(x_k,y_k) + b(x_k,y_k) \right]
\end{equation}

For persistent data gaps, multi-temporal integration combines estimates from multiple reference times $\{t_i \mid i = 1, \dots, n\}$ through weighted averaging:

\begin{align}
\overline{S_j}(x,y) &= \sum_{p=l_1}^{l_m} wt_p \cdot S_j^p(x,y) \\
wt_p &= \frac{1/DT_p}{\sum_{p=1}^M (1/DT_p)}
\end{align}

where $DT_p$ quantifies temporal variance between reference time $t_p$ and target time $t_j$, giving greater weight to temporally stable estimates.

\subsubsection{Implementation Details}

The fusion process begins by identifying similar grid cells that satisfy both spatial and consistency thresholds: 
\begin{align}
|S_i(x,y) - S_i(x_k,y_k)| &< 1.5\, \mu g/m^3 \quad \text{(spatial similarity)} \\
|S_i(x_k,y_k) - G_i(x_k,y_k)| &< 2.0\, \mu g/m^3 \quad \text{(data consistency)}
\end{align}

These empirically optimized thresholds ensure reliable neighborhood selection while accounting for measurement uncertainties.

Linear coefficients $a$ and $b$ are derived through weighted least-squares regression between ground observations $G_i$ and $G_j$ at analogous locations, using Huber loss for robustness against outliers. The weights $w(x_k, y_k)$ combine spatial proximity and concentration similarity:

\begin{align}
D(x_k,y_k) &= |S_i(x,y) - S_i(x_k,y_k)| \\
w(x_k,y_k) &= \frac{1/D(x_k,y_k)}{\sum_{k=1}^N (1/D(x_k,y_k))}
\end{align}

Computational efficiency is achieved through parallel processing across grid cells and spatial indexing for rapid neighborhood searches. The implementation handles large datasets via memory-mapping techniques, maintaining the native $0.05^\circ$ grid resolution while demonstrating a 14\% improvement in RMSE for SO\textsubscript{2} and 9\% for NO\textsubscript{2} compared to conventional interpolation methods (Section~\ref{sec:performance}, Table~\ref{tab:metrics1}).

\subsection{Vision Transformer Architecture for Pollutant Prediction}

\begin{figure*}[t]
    \centering
    \includegraphics[width=0.95\textwidth]{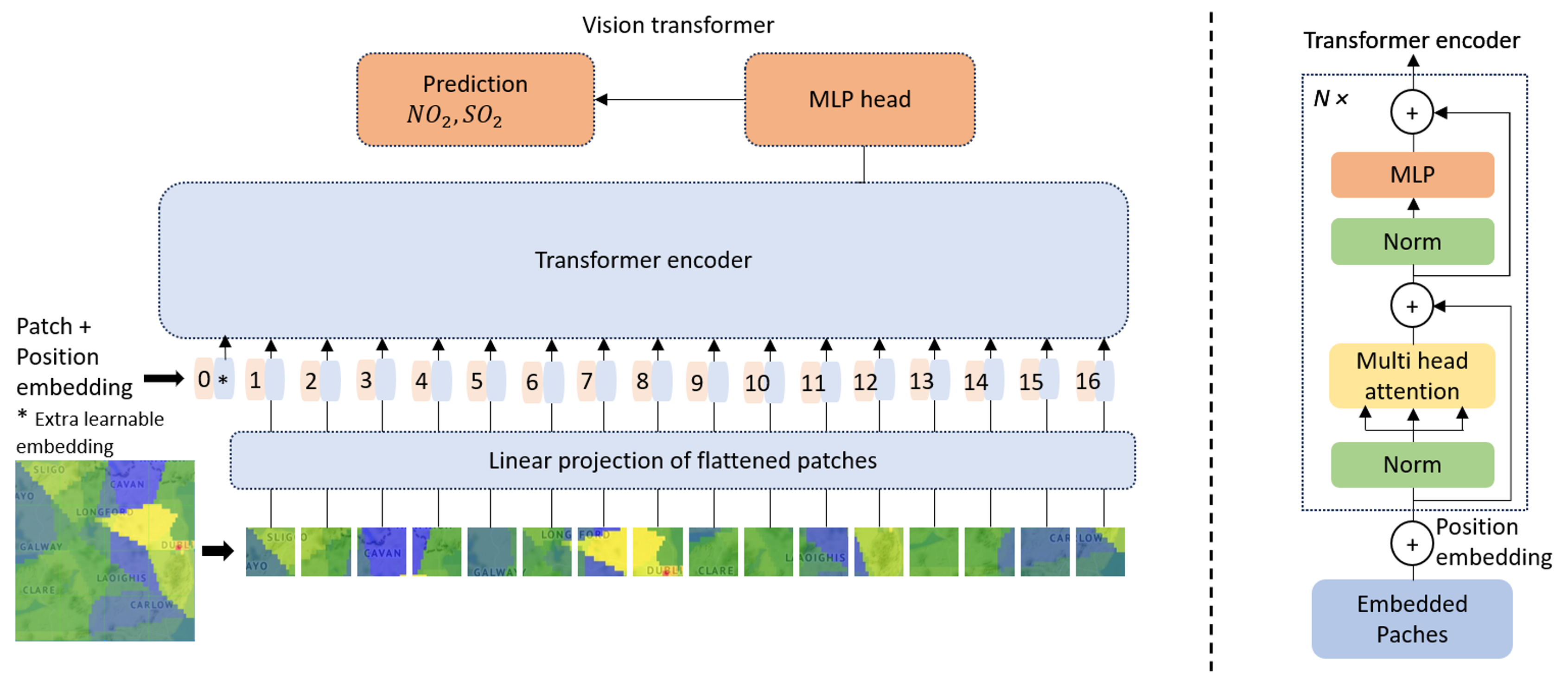}
    \caption{Architecture of the Vision Transformer (ViT) for NO\textsubscript{2} and SO\textsubscript{2} concentration prediction, showing: (1) patch embedding and positional encoding, (2) multi-head self-attention layers, and (3) MLP projection head.}
    \label{fig:transformer}
\end{figure*}

The Vision Transformer (ViT) architecture, adapted from the original transformer framework developed for natural language processing~\cite{gomez2017attention}, offers significant advantages for processing spatial-temporal pollution data. As shown in Fig.~\ref{fig:transformer}, our ViT implementation transforms 2D concentration maps of NO\textsubscript{2} and SO\textsubscript{2} into sequences of patch embeddings that capture both local and global atmospheric patterns.

\subsubsection{Patch Processing and Embedding}

The input concentration map \(X \in \mathbb{R}^{h \times w \times c}\) (where \(c\) denotes channels) is divided into \(n\) flattened 2D patches \(X_p \in \mathbb{R}^{n \times (p^2c)}\), with each patch of size \(p \times p\) pixels. The sequence length \(n = hw/p^2\) is determined by the original image dimensions and patch size. Each patch undergoes linear projection into a \(d\)-dimensional embedding space:

\begin{equation}
    E^{ij} = W_p X_p^{ij} + b_p
\end{equation}

where \(E^{ij}\) represents the embedding vector for patch \(j\) in sample \(i\), with learnable parameters \(W_p \in \mathbb{R}^{d \times (p^2c)}\) and bias \(b_p\). Positional encodings are added to preserve spatial information, crucial for maintaining the geographic relationships between atmospheric measurements.

\subsubsection{Self-Attention Mechanism}

The core innovation of ViT lies in its self-attention mechanism, which computes relationships between all patches simultaneously. For an input sequence, the model generates query (\(Q\)), key (\(K\)), and value (\(V\)) matrices through learned linear transformations. The attention weights are computed as:

\begin{equation}
    \text{Attention}(Q,K,V) = \text{softmax}\left(\frac{QK^T}{\sqrt{d_k}}\right)V
\end{equation}

This process occurs in four stages. First, score calculation is performed by computing pairwise patch similarities via \(QK^T\). Next, normalization is applied by scaling the scores with \(\sqrt{d_k}\) to ensure stable gradients. This is followed by probability mapping, where a softmax function is used to generate attention weights. Finally, value weighting combines the values according to these attention weights, allowing the model to capture both local and global dependencies within the data.

\subsubsection{Multi-Head Attention and Network Architecture}

To capture diverse relationships, we employ multi-head attention:

\begin{align}
    \text{MultiHead}(Q,K,V) &= \text{Concat}(\text{head}_1,...,\text{head}_h)W^\circ \\
    \text{head}_i &= \text{Attention}(QW_i^Q, KW_i^K, VW_i^V)
\end{align}

where \(h=8\) parallel attention heads project inputs into different subspaces using learned matrices \(W_i^Q, W_i^K, W_i^V \in \mathbb{R}^{d \times d_k}\), followed by concatenation and projection via \(W^\circ \in \mathbb{R}^{hd_v \times d}\).

The transformer encoder alternates between multi-head attention layers and multilayer perceptrons (MLPs) with Gaussian Error Linear Unit (GELU) activation:

\begin{equation}
    \text{MLP}(X_p) = \text{GELU}(W_1X_p)W_2
\end{equation}

where \(W_1 \in \mathbb{R}^{d \times d_{ff}}\) and \(W_2 \in \mathbb{R}^{d_{ff} \times d}\) form the two-layer feedforward network. This architecture, with 12 encoder blocks and 64 embedding dimensions, effectively models both local pollutant variations and regional atmospheric patterns.
%-------------------------------------------------------------------------------------------

\section{Experimental Results and Discussion} \label{sec:result}

We evaluate PollutionNet's performance against four baseline models: CNN, linear regression (LR), XGBoost (XGB), and LightGBM (LGBM). The comparative analysis demonstrates our model's superior capability in predicting NO\textsubscript{2} and SO\textsubscript{2} concentrations.

\subsection{Model Configuration and Training}

The dataset comprises 485 samples for each pollutant, split using five-fold cross-validation (80\% training, 20\% validation). Table~\ref{tab:hyperparameter} summarizes the optimal hyperparameters identified for each model:

\begin{table}[!ht]
\caption{Optimal hyperparameter configurations for all models}
\label{tab:hyperparameter}
\begin{tabularx}{\linewidth}{lXXXXX}
\toprule
\textbf{Parameter} & \textbf{PollutionNet} & \textbf{CNN} & \textbf{LR} & \textbf{XGB} & \textbf{LGBM} \\
\midrule
Epochs & 30 & 30 & -- & -- & -- \\
Learning Rate & 0.01 & 0.01 & -- & 0.1 & 0.1 \\
Optimizer & Adam & Adam & -- & -- & -- \\
Activation & GELU & ReLU & -- & -- & -- \\
Batch Size & 8 & 8 & -- & -- & -- \\
Patch Size & 16 & -- & -- & -- & -- \\
Embedding Dim & 64 & -- & -- & -- & -- \\
Attention Heads & 8 & -- & -- & -- & -- \\
Transformer Blocks & 12 & -- & -- & -- & -- \\
Kernel Size & -- & 3$\times$3 & -- & -- & -- \\
Regularization & -- & -- & -- & $\gamma$=0, $\alpha$=0, $\lambda$=1 & $\alpha$=0, $\lambda$=0 \\
\bottomrule
\end{tabularx}
\end{table}

\subsection{Performance Evaluation of PollutionNet}
\label{sec:performance}

\begin{figure}[ht!]
    \centering
    \includegraphics[width=0.30\textwidth]{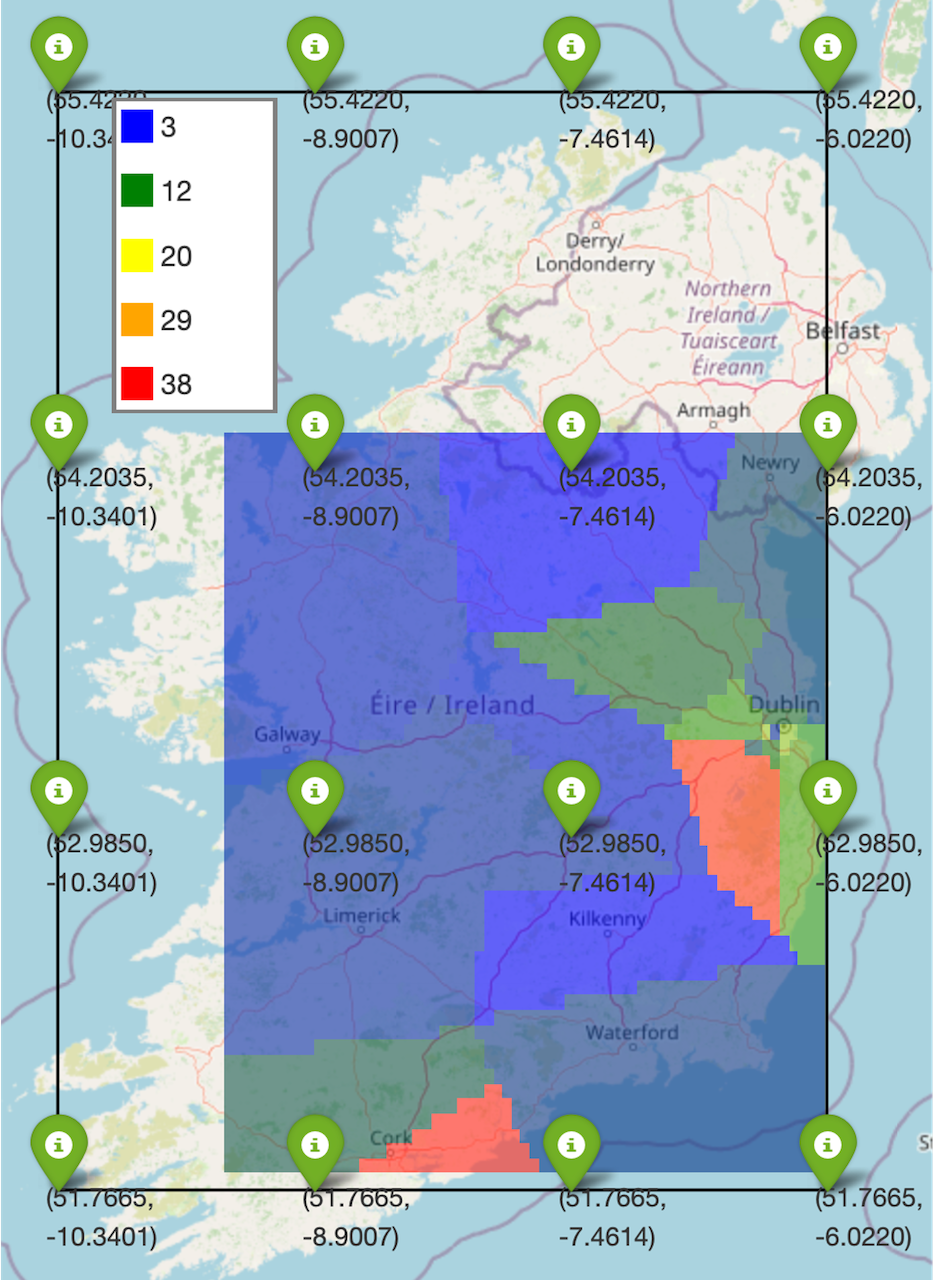}
     %\vspace{5pt}
     \includegraphics[width=0.30\textwidth]{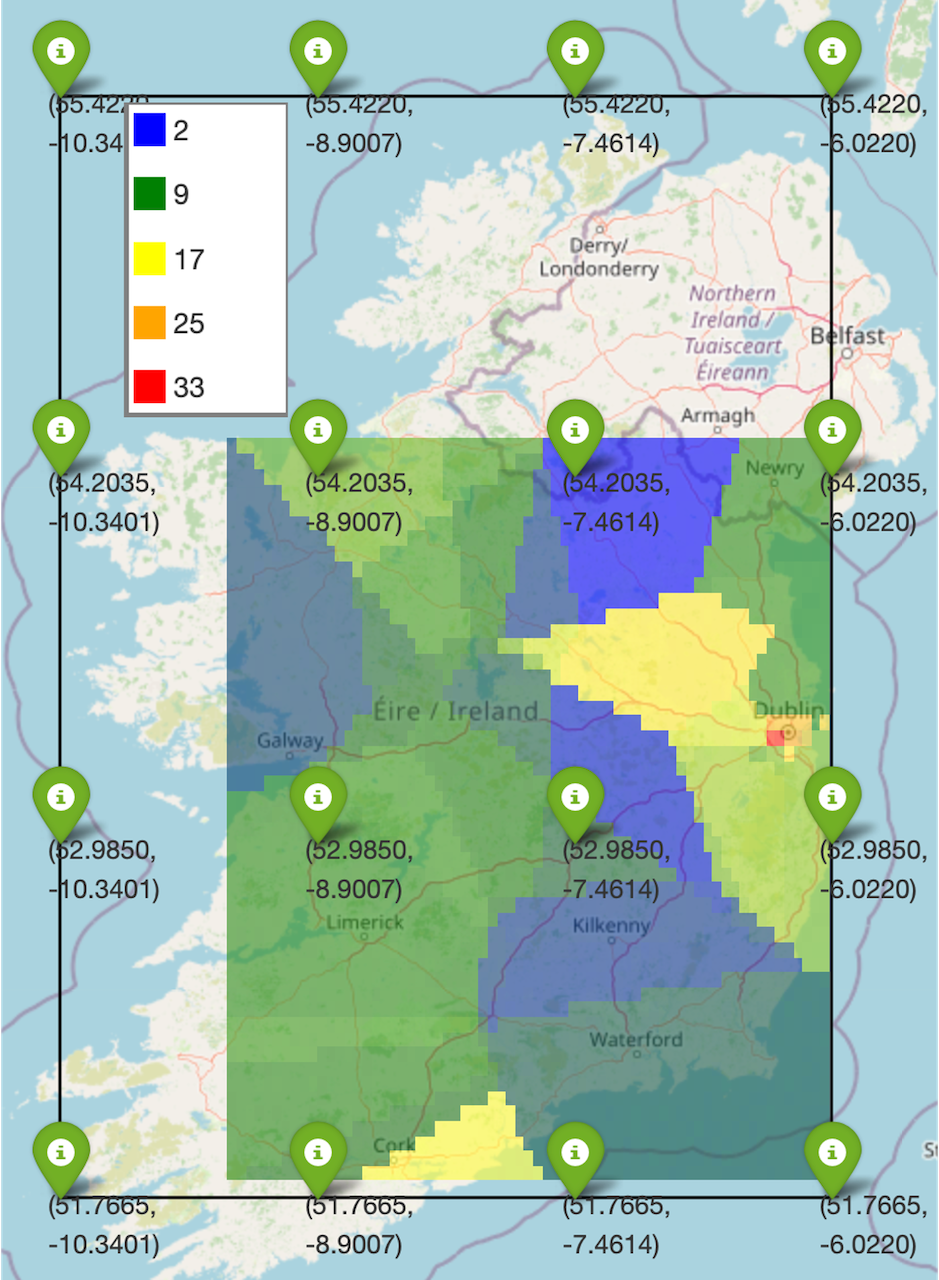}
     %\vspace{5pt}
     \includegraphics[width=0.30\textwidth]{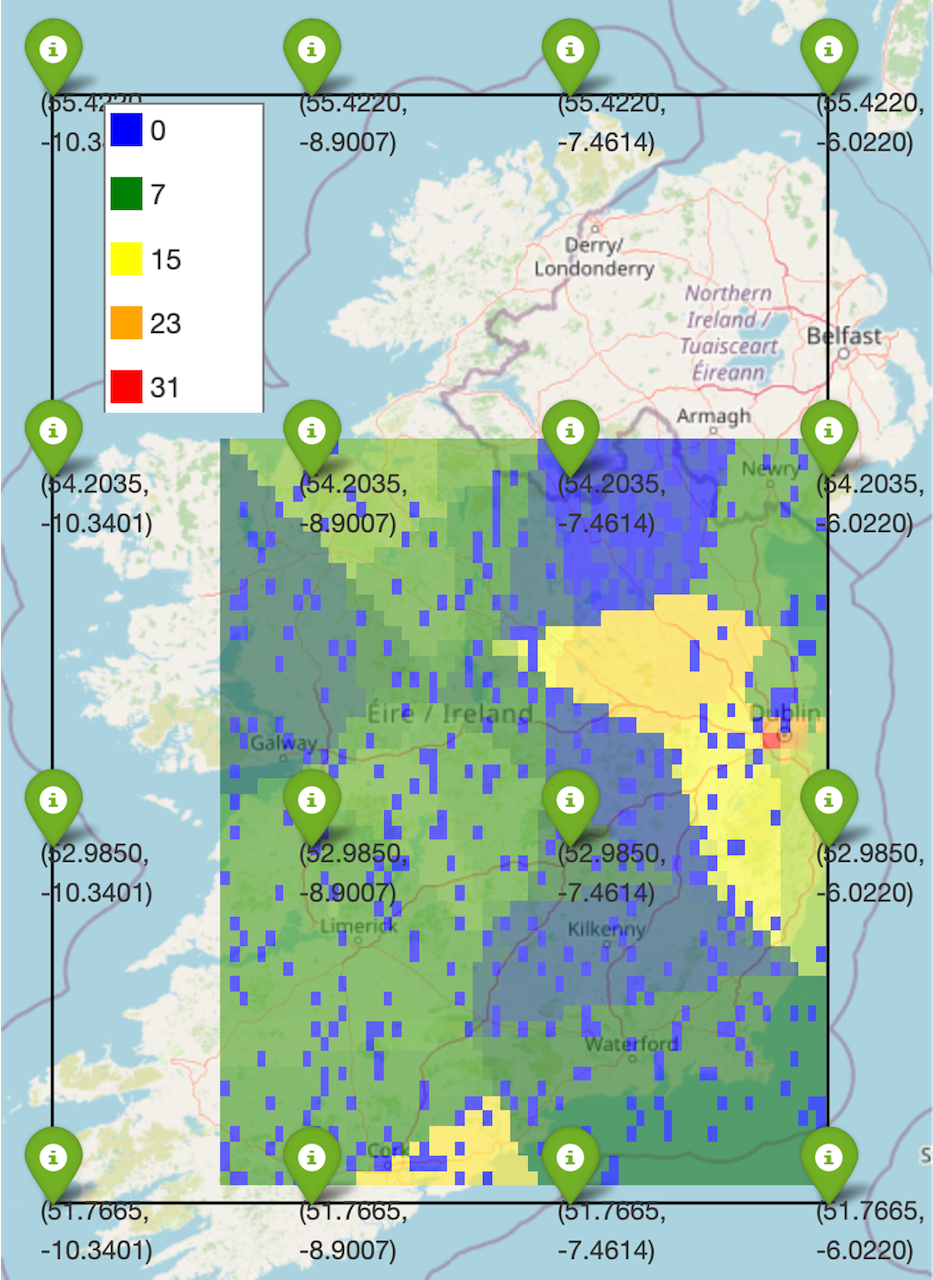}
     \\
     \makebox[0.30\textwidth][c]{(a) Ground True}
     \makebox[0.30\textwidth][c]{(b) PollutionNet}
     \makebox[0.30\textwidth][c]{(c) CNN} 
     \\
    \includegraphics[width=0.30\textwidth]{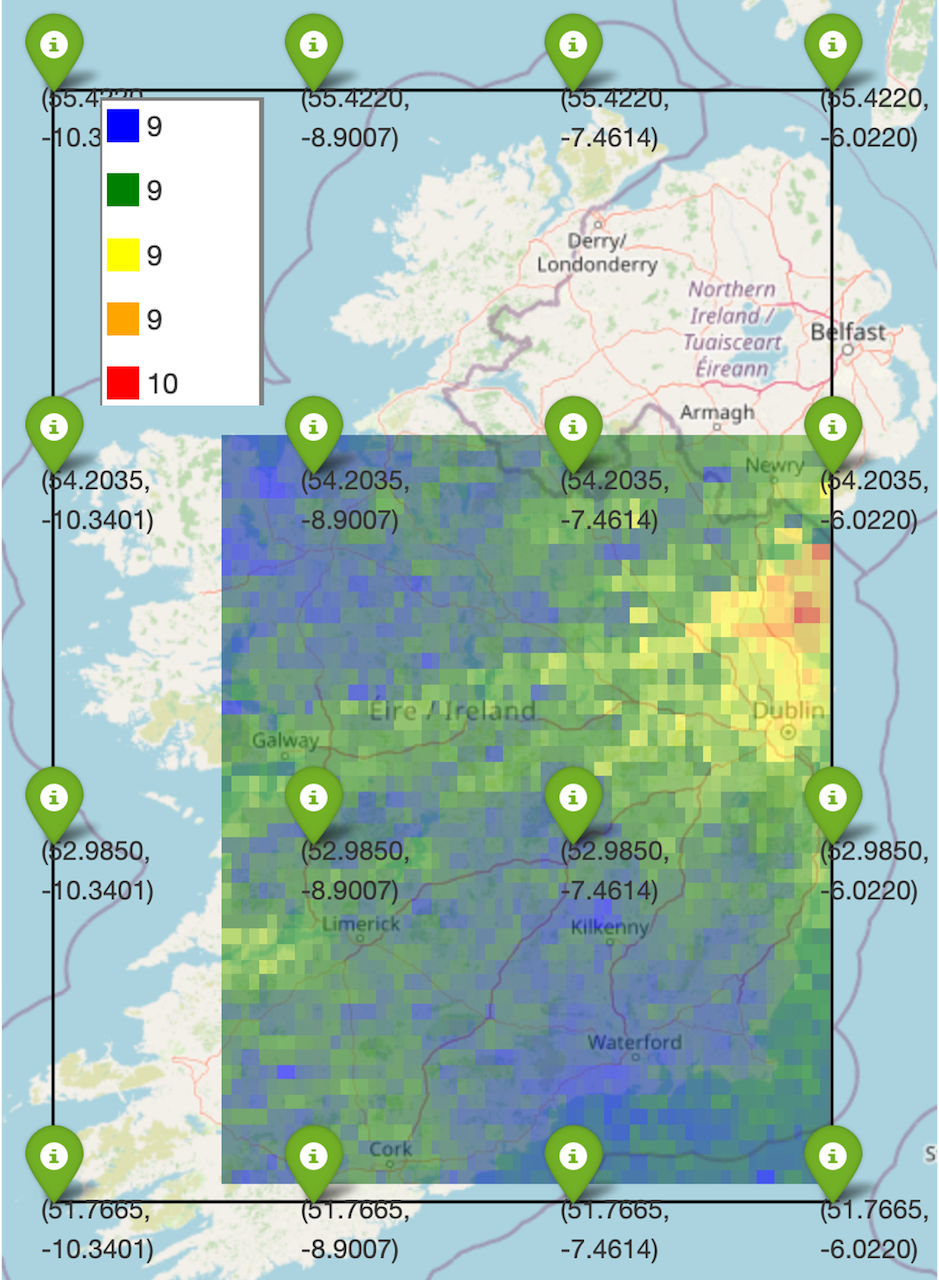}
     %\vspace{5pt}
     \includegraphics[width=0.30\textwidth]{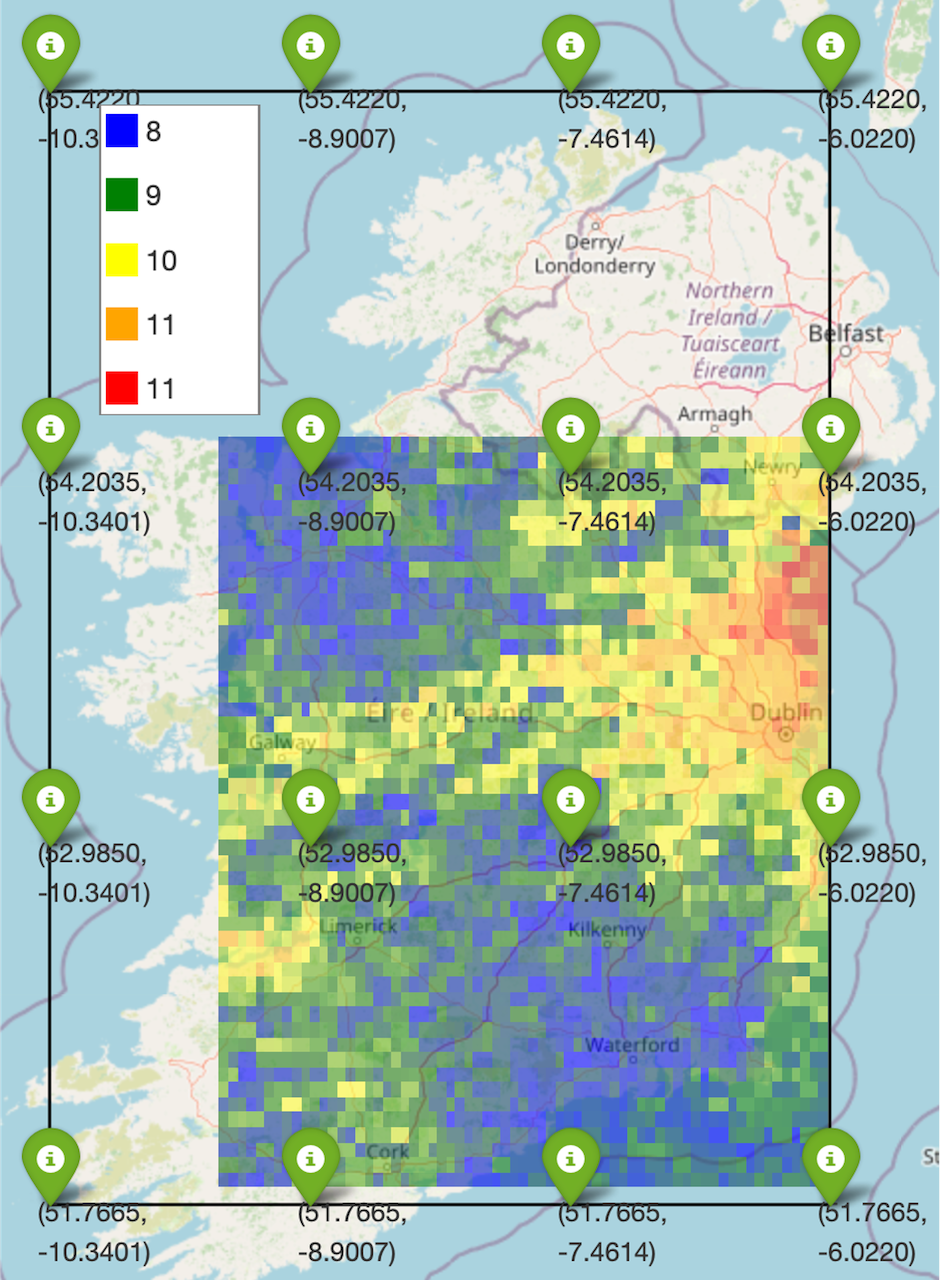}
     \includegraphics[width=0.30\textwidth]{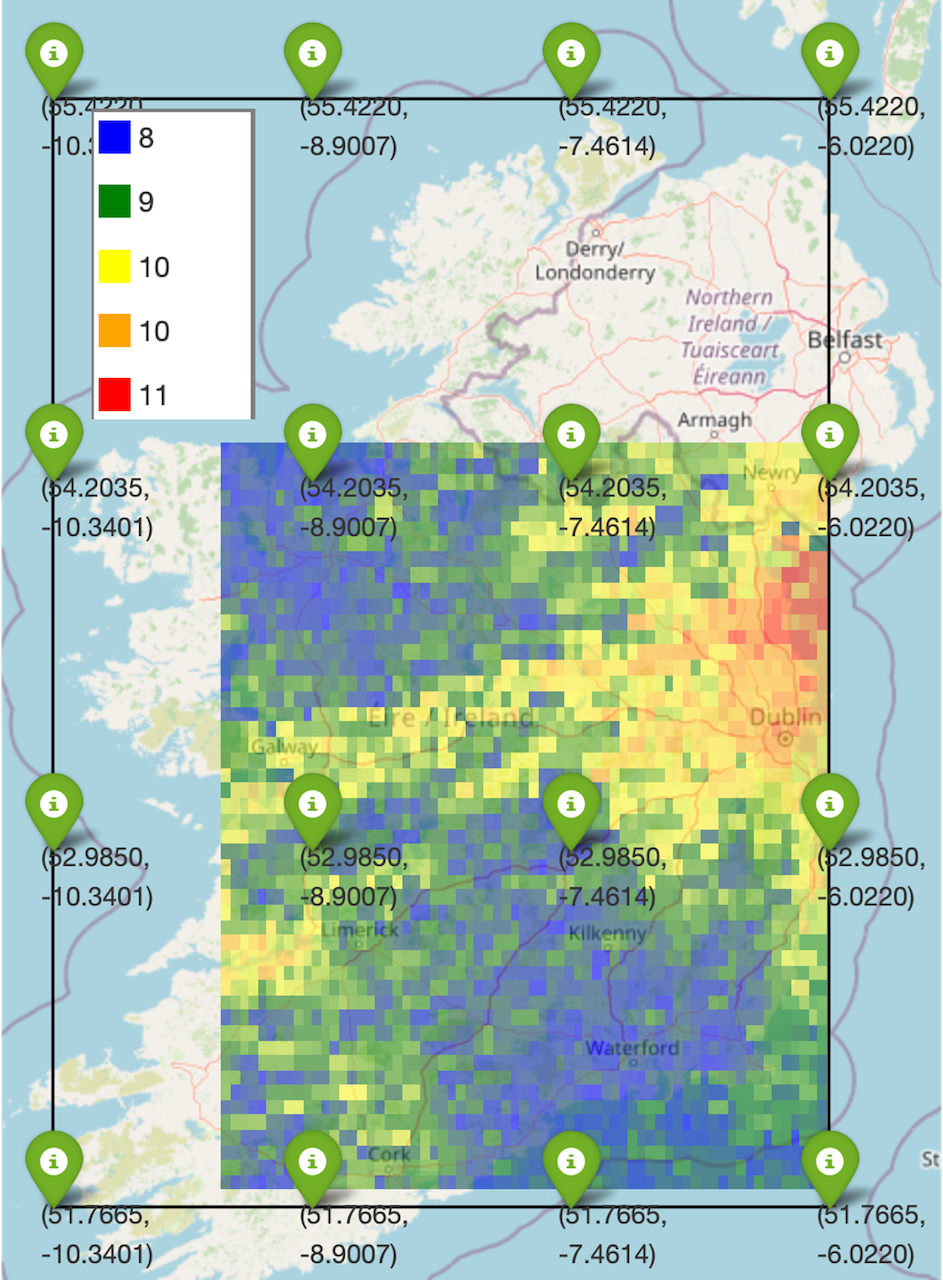}
     \\
     \makebox[0.30\textwidth][c]{(d) LR}
     \makebox[0.30\textwidth][c]{(e) XGBoost} 
     \makebox[0.30\textwidth][c]{(f) LGBM} 
     
    \caption{Depicting the daily average of ground-level concentration vs. the predicted NO$_2$ concentration from various models: PollutionNet, CNN, linear regression, XGBoost, and LGBM.}
    \label{fig:pollution_pred1}
\end{figure}

\begin{figure}[ht!]
    \centering
     \includegraphics[width=0.30\textwidth]{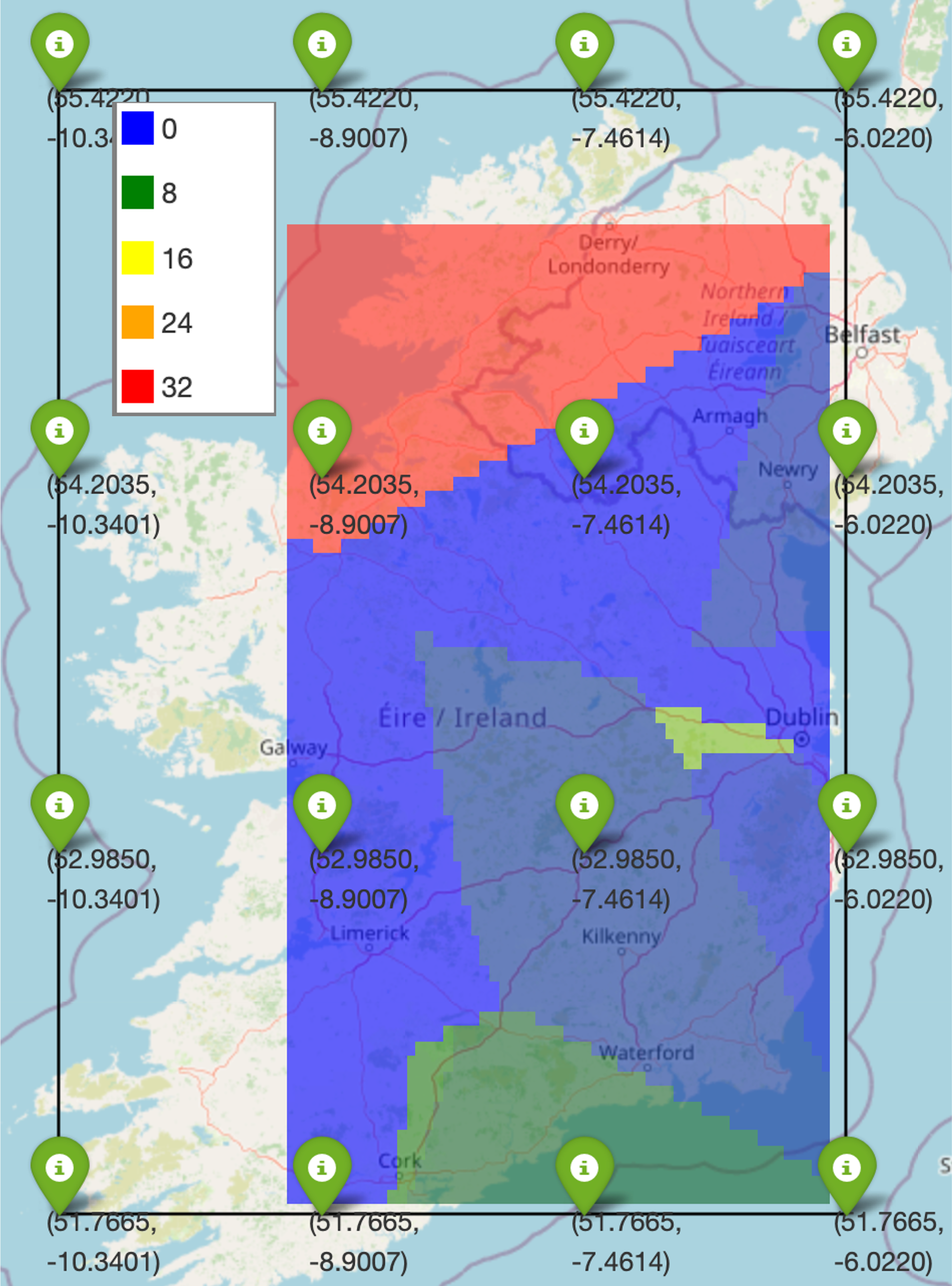}
     %\vspace{5pt}
     \includegraphics[width=0.30\textwidth]{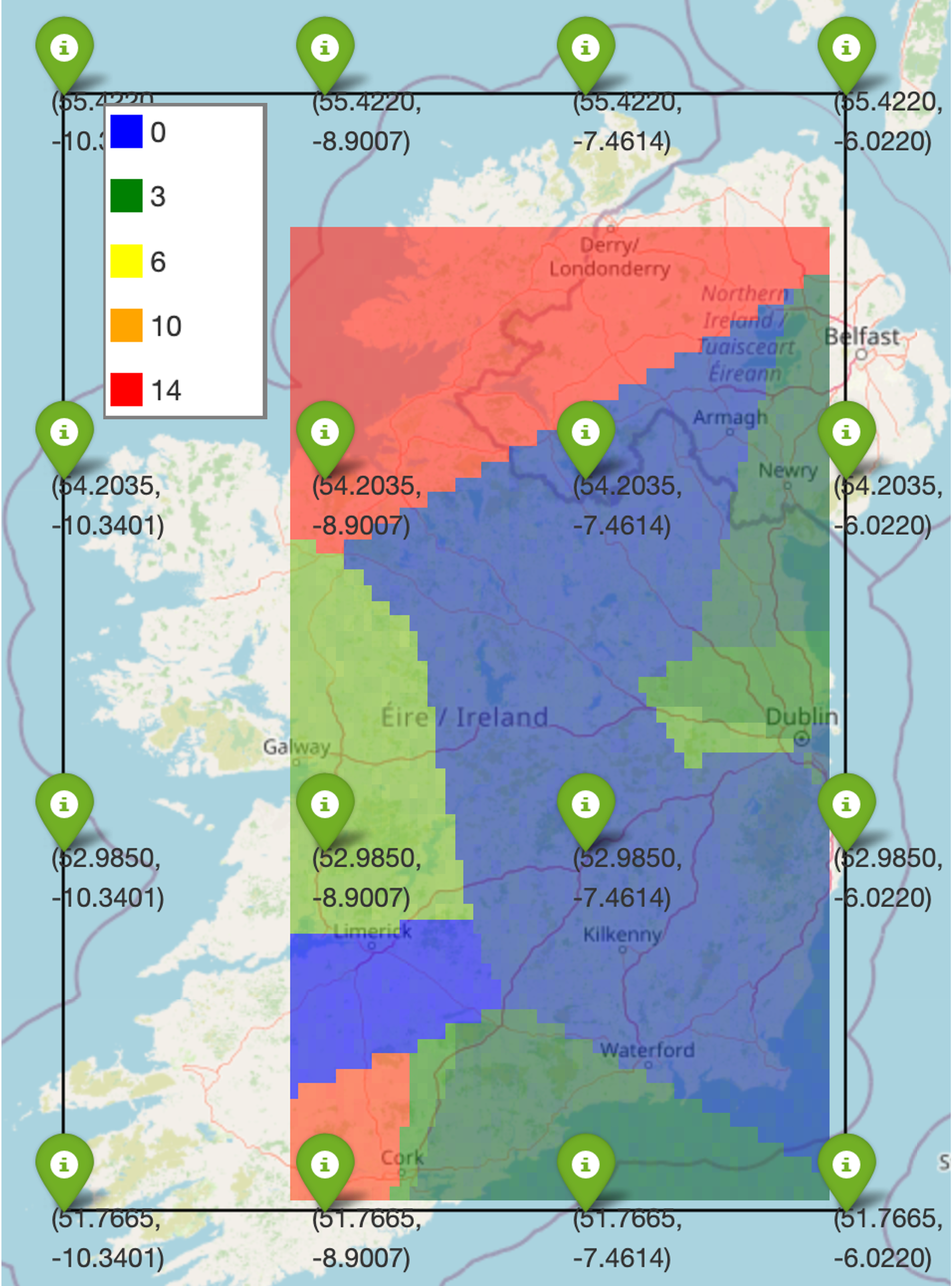}
     %\vspace{5pt}
     \includegraphics[width=0.30\textwidth]{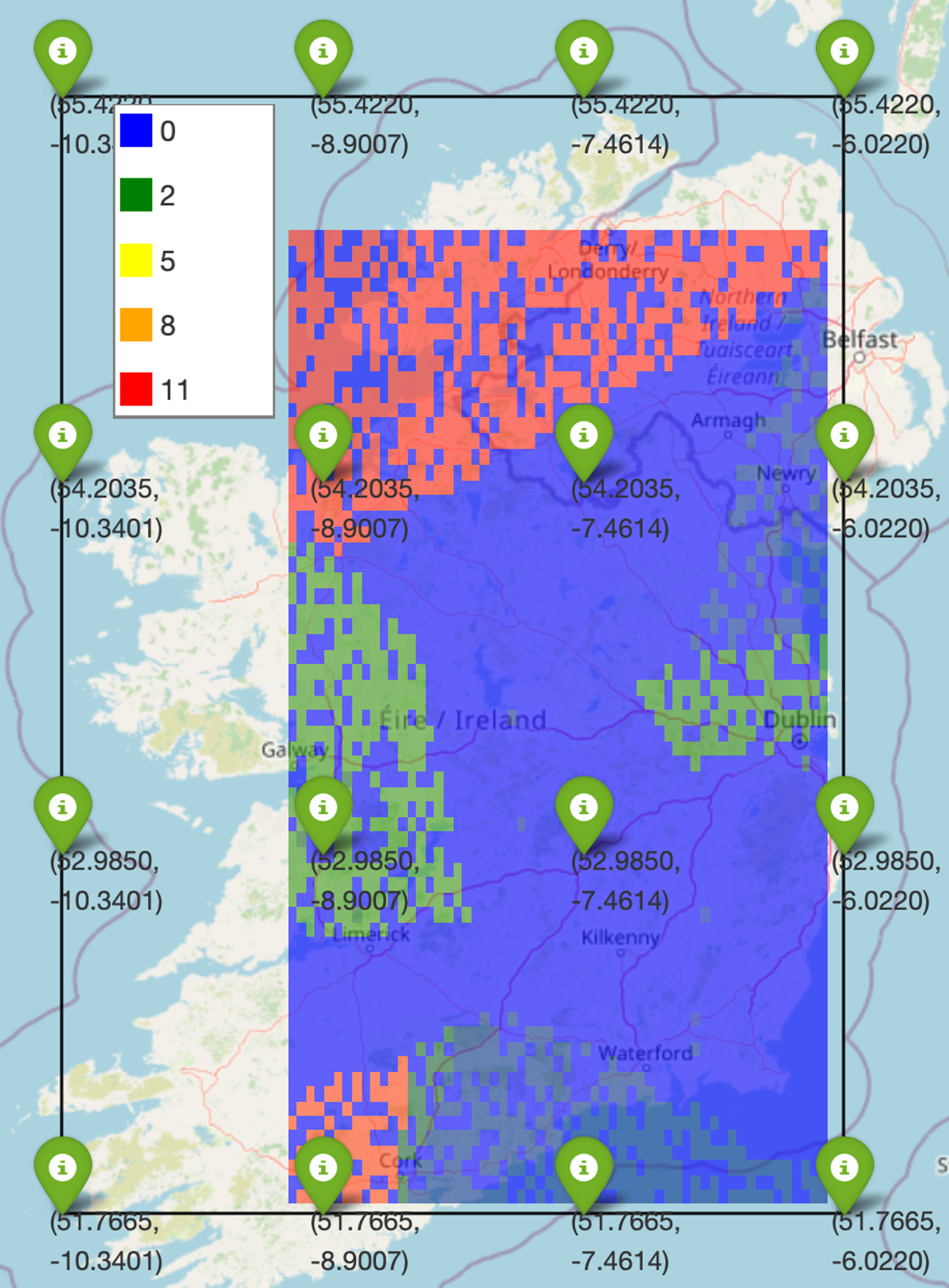}
     \\
     \makebox[0.30\textwidth][c]{(a) Ground True}
     \makebox[0.30\textwidth][c]{(b) PollutionNet} 
     \makebox[0.30\textwidth][c]{(c) CNN} 
     \\
    \includegraphics[width=0.30\textwidth]{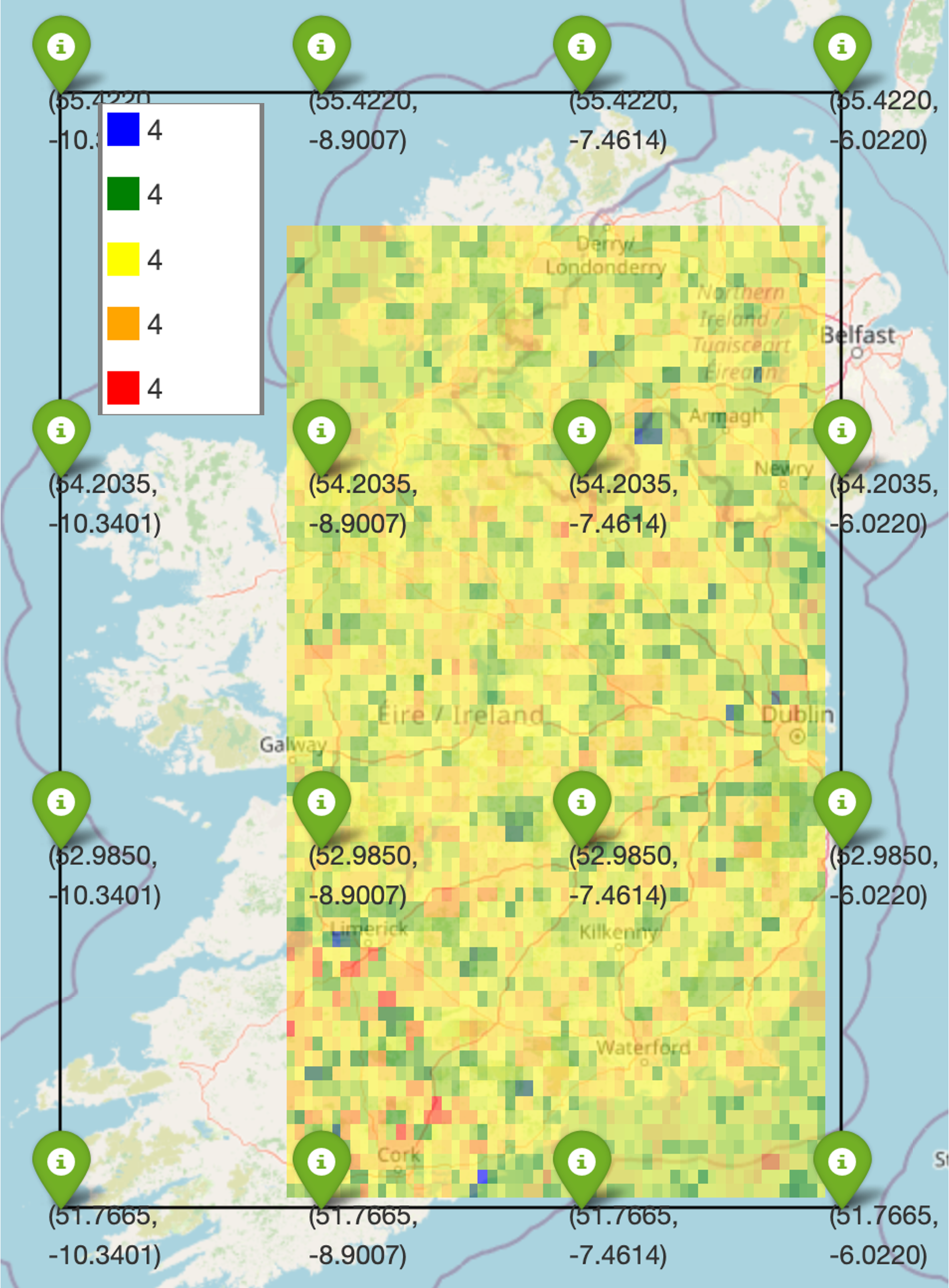}
     %\vspace{5pt}
     \includegraphics[width=0.30\textwidth]{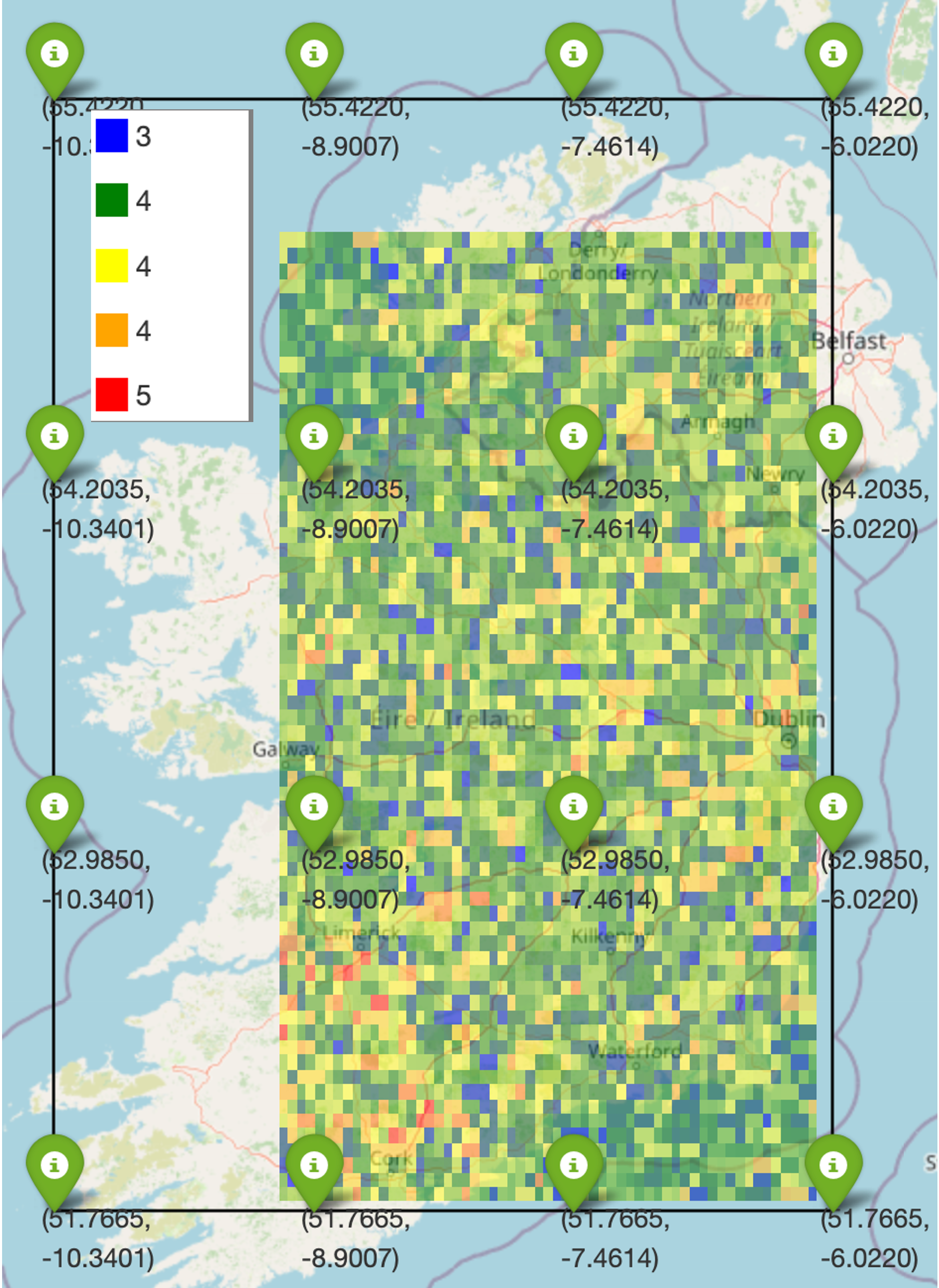}
    \includegraphics[width=0.30\textwidth]{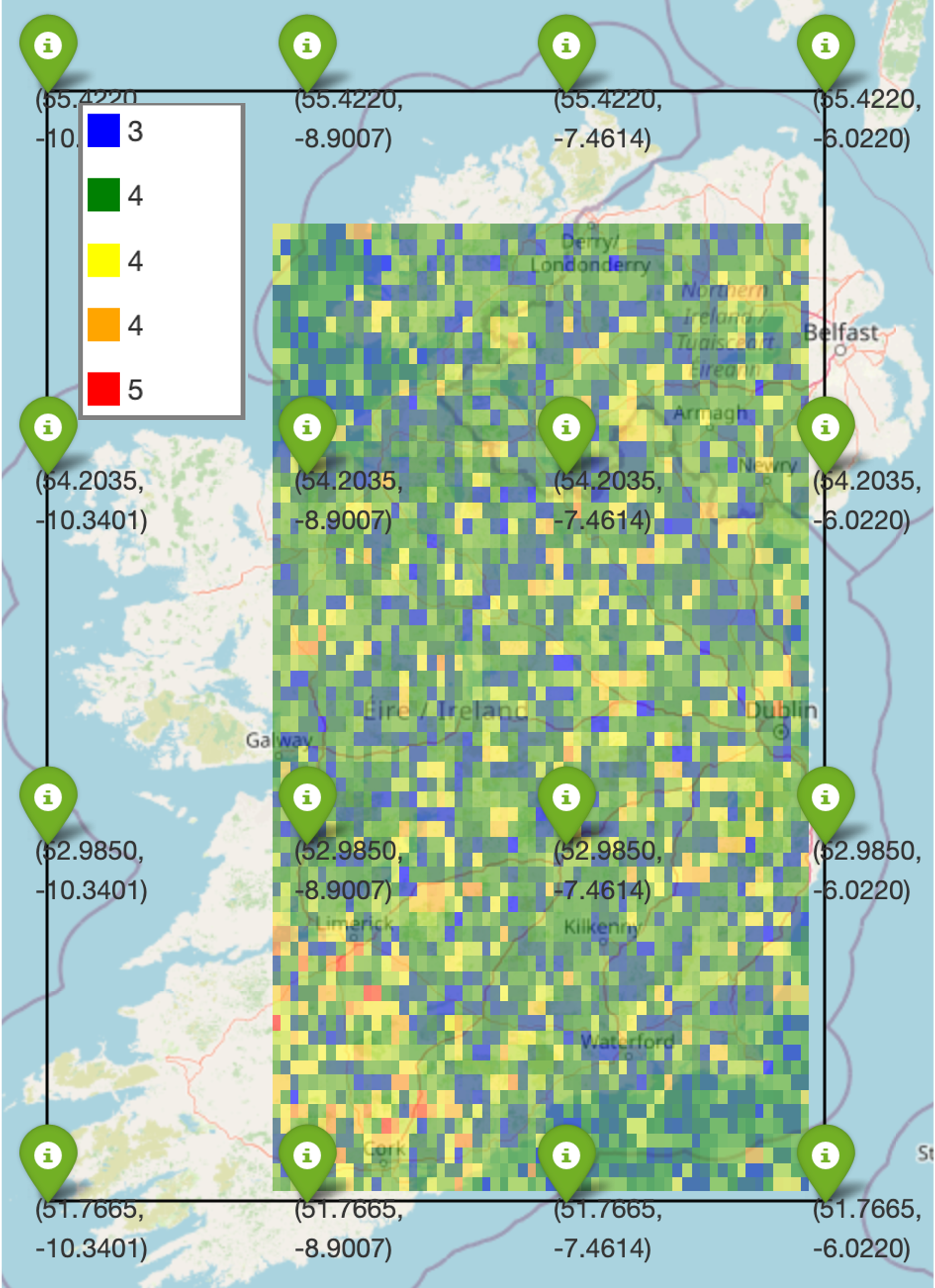}
     \\
     \makebox[0.30\textwidth][c]{(d) LR}
     \makebox[0.30\textwidth][c]{(e) XGBoost}
     \makebox[0.30\textwidth][c]{(f) LGBM}

    \caption{Depicting the daily average of ground-level concentration vs. the predicted SO$_2$ concentration from various models: PollutionNet, CNN, linear regression, XGBoost, and LGBM.}
    \label{fig:pollution_pred2}
\end{figure}

The proposed PollutionNet framework demonstrates superior performance in predicting surface-level concentrations of NO\textsubscript{2} and SO\textsubscript{2} compared to conventional models, including CNN, linear regression (LR), XGBoost, and LGBM. Figs.~\ref{fig:pollution_pred1} and~\ref{fig:pollution_pred2} present a comparative visualization of daily average pollutant concentrations, contrasting ground-truth measurements with model predictions. PollutionNet effectively captures localized spatial patterns in NO\textsubscript{2} and SO\textsubscript{2} distributions, whereas other models exhibit weaker correlations and fail to reproduce the fine-scale variations observed in the actual data.

\begin{table}[!ht]
 \centering
 \caption{Performance comparison of PollutionNet with CNN, Linear Regression, XGBoost, and LGBM using RMSE, MAE, and Pearson's Correlation Coefficient (R$^2$).}
%\resizebox{\linewidth}{!}{%
%\begin{tabular}{@{}l*{13}{c}@{}}
\begin{tabular}{cc|ll|ll|ll|ll|ll}
\toprule
 &\textbf{Models} & \multicolumn{2}{c|}{\textbf{Proposed}} & \multicolumn{2}{c|}{\textbf{CNN}} & \multicolumn{2}{c|}{\textbf{Linear}} & \multicolumn{2}{c|}{\textbf{XGB}} & \multicolumn{2}{c}{\textbf{LGBM}} \\ 
 \cmidrule(lr){3-4} \cmidrule(lr){5-6} \cmidrule(lr){7-8} \cmidrule(lr){9-10} \cmidrule(lr){11-12} &  & \textbf{NO$_2$} & \textbf{SO$_2$}  & \textbf{NO$_2$} & \textbf{SO$_2$}  & \textbf{NO$_2$} & \textbf{SO$_2$} & \textbf{NO$_2$} & \textbf{SO$_2$}  & \textbf{NO$_2$} & \textbf{SO$_2$} \\ \midrule

\multirow{6}{*}{\rotatebox[origin=c]{90}{\textbf{RMSE}}} & Fold1 & 7.86 & 4.91 & 8.38 & 6.33 & 8.95& 6.83 & 8.98 & 6.84  & 8.97 & 6.23 \\ 
& Fold2 & 6.51 & 4.34 & 6.97 & 5.54 & 7.55 & 6.24 & 7.50 & 6.24 & 7.74 & 6.82 \\  
& Fold3 & 7.09 & 4.71 & 7.59 & 5.13 & 8.22 & 6.68 & 8.11 & 6.72 & 8.10 & 6.70  \\ 
& Fold4 & 6.45 & 4.92 & 8.00 & 5.41 & 7.62 & 6.79 & 7.59 & 6.83 & 7.59 & 6.81 \\  
& Fold5 & 6.56 & 3.57 & 7.30 & 3.86 & 7.63 & 5.44 & 7.60 & 5.42 & 7.59 & 5.41 \\ 
\midrule
& \textbf{Avg} &\textbf{6.89} & \textbf{4.49} & \textbf{7.65} &\textbf{5.25} & \textbf{8.00}& \textbf{6.39}  & \textbf{7.96} &\textbf{6.41} & \textbf{7.95} & \textbf{6.39} \\ \midrule
\multirow{6}{*}{\rotatebox[origin=c]{90}{\textbf{MAE}}} & Fold1 & 5.15 & 3.02 & 5.37 & 3.89 & 6.04 & 4.72 & 5.93 & 4.71 & 5.91 & 4.71 \\ 
& Fold2 & 5.74 & 3.18 & 6.45 & 4.33 & 6.77 & 5.06 & 6.81 & 5.06 & 6.80 & 5.06 \\  
& Fold3 & 5.50 & 3.07 & 6.04 & 3.29 & 6.46 & 4.84 & 6.42 & 4.88 & 6.42 & 4.88  \\ 
& Fold4 & 4.88 & 3.18 & 6.24 & 3.58 & 5.91 & 4.93 & 5.86 & 4.96 & 5.85 & 4.95 \\  
& Fold5 & 5.22 & 2.83 & 5.81 & 2.79 & 6.15 & 4.39 & 6.10 & 4.37 & 6.10 & 4.37 \\  
\midrule
& \textbf{Avg} & \textbf{5.31} & \textbf{3.06} & \textbf{5.98} & \textbf{3.58} & \textbf{6.27} &\textbf{4.79} & \textbf{6.22} &\textbf{4.80} & \textbf{6.22} &\textbf{4.79} \\ \midrule
\multirow{6}{*}{\rotatebox[origin=c]{90}{\textbf{R$^2$}}} & Fold1 & 0.63 & 0.77& 0.52&0.46 &0.08 &0.007 &0.03 &-0.02 & 0.03&-0.02 \\ 
& Fold2 & 0.64 & 0.78 & 0.52 & 0.35 & 0.07 & 0.003 & 0.05 & -0.01 & 0.06 & -0.01\\  
& Fold3 & 0.66 & 0.77 & 0.54 & 0.72 & 0.06 & -0.01 & 0.03 & -0.03 & 0.04 & -0.03 \\  
& Fold4 & 0.64 & 0.78 & 0.44 & 0.68 & 0.08 & 0.004 & 0.04 & -0.01 & 0.04 & -0.01 \\  
& Fold5 & 0.64 & 0.76 & 0.49 & 0.72 & 0.08 & -0.004 & 0.04 & -0.01 & 0.04 & -0.01 \\ 
\midrule
& \textbf{Avg} &\textbf{0.64} &\textbf{0.77} &\textbf{0.50} &\textbf{0.58} &\textbf{0.07} &\textbf{-0.001} & \textbf{0.03}&\textbf{-0.01}  &\textbf{0.04} &\textbf{-0.01} \\ \bottomrule
\end{tabular}
%}
\label{tab:metrics1}
\end{table}

Quantitative validation, as summarized in Table~\ref{tab:metrics1}, reinforces these observations. PollutionNet achieves an RMSE of 6.89~\textmu g/m\textsuperscript{3} and 4.49~\textmu g/m\textsuperscript{3} for NO\textsubscript{2} and SO\textsubscript{2}, respectively, outperforming all benchmarked models. Similarly, the MAE values (5.31~\textmu g/m\textsuperscript{3} for NO\textsubscript{2}, 3.06~\textmu g/m\textsuperscript{3} for SO\textsubscript{2}) indicate higher precision compared to competing approaches. Notably, Pearson’s correlation coefficient ($R^2$) further highlights PollutionNet’s robustness, yielding 0.64 for NO\textsubscript{2} and 0.77 for SO\textsubscript{2}, significantly higher than those of CNN, LR, XGBoost, and LGBM.

A key advantage of PollutionNet is its consistency across cross-validation folds, with minimal performance fluctuations. The framework reduces RMSE by 9\% (NO\textsubscript{2}) and 14\% (SO\textsubscript{2}), MAE by 11\% (NO\textsubscript{2}) and 14\% (SO\textsubscript{2}), and improves $R^2$ by 28\% (NO\textsubscript{2}) and 32\% (SO\textsubscript{2}) compared to the next-best model. These results underscore PollutionNet’s ability to generalize across different data partitions while maintaining high predictive accuracy.

\subsection{Temporal Analysis and Stability}

\begin{figure}[!ht]
    \centering
    \includegraphics[width=.95\textwidth]{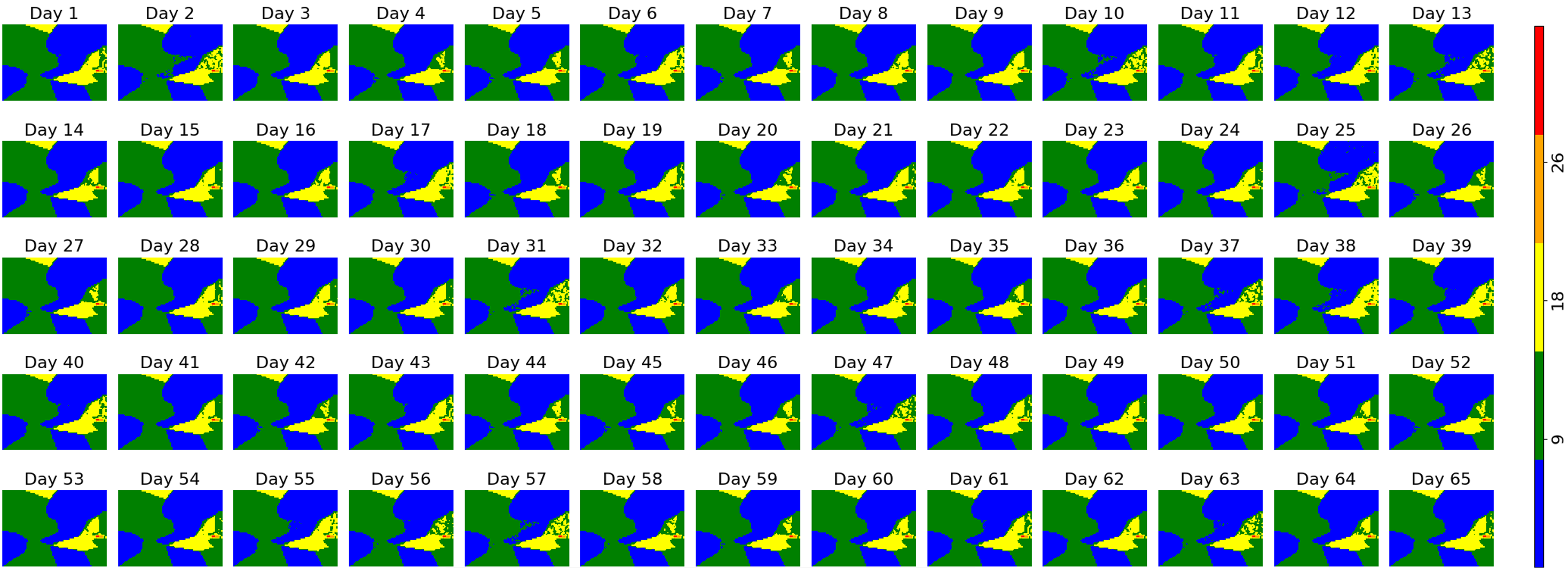}
    \caption{Daily time-series (March to April 2021) heatmap of PollutionNet for predicting NO$_2$ concentrations over the Ireland region.}
    \label{fig:temporal}
\end{figure}

To assess PollutionNet’s temporal reliability, we analyzed daily NO\textsubscript{2} predictions over a two-month period (March–April 2021) in Ireland (Fig.~\ref{fig:temporal}). The model accurately captures minor fluctuations in pollutant concentrations, suggesting stable emission patterns during this period. The absence of significant temporal anomalies indicates that PollutionNet reliably tracks NO\textsubscript{2} trends without overfitting to short-term variations.

\subsection{Comparative Performance via Joint Distribution Analysis}

\begin{figure}[!ht]
    \centering
    \footnotesize
    \includegraphics[width=0.19\textwidth]{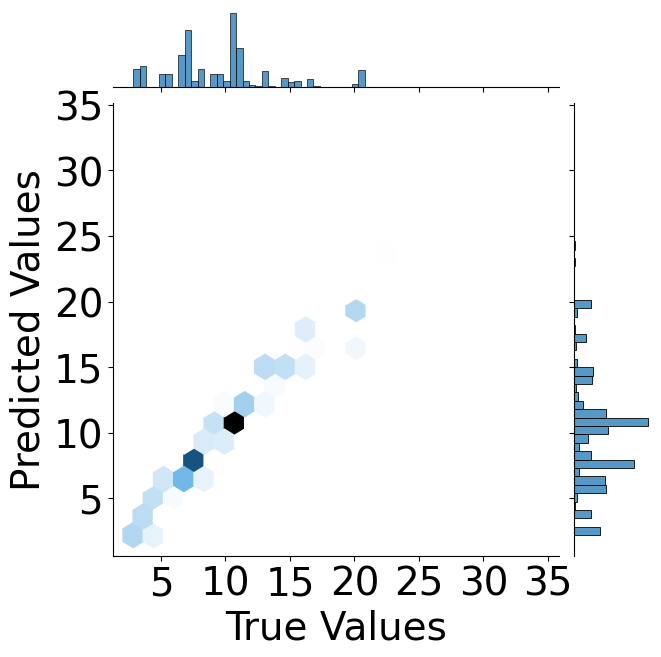}
     %\vspace{5pt}
     \includegraphics[width=0.19\textwidth]{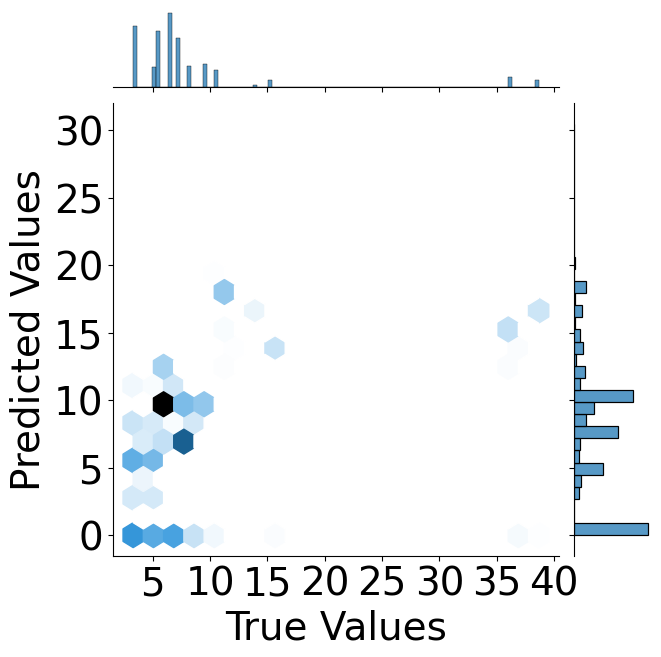}
     %\vspace{5pt}
     \includegraphics[width=0.19\textwidth]{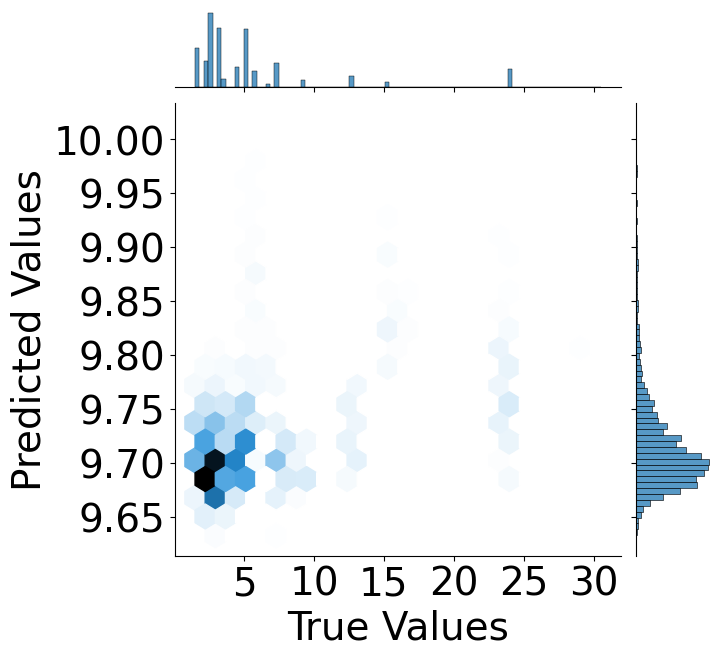}
     \includegraphics[width=0.19\textwidth]{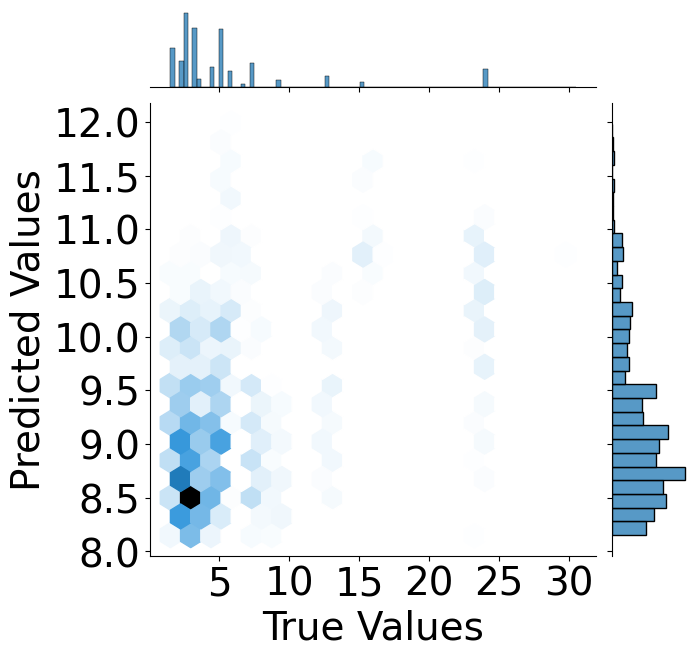}
     %\vspace{5pt}
     \includegraphics[width=0.19\textwidth]{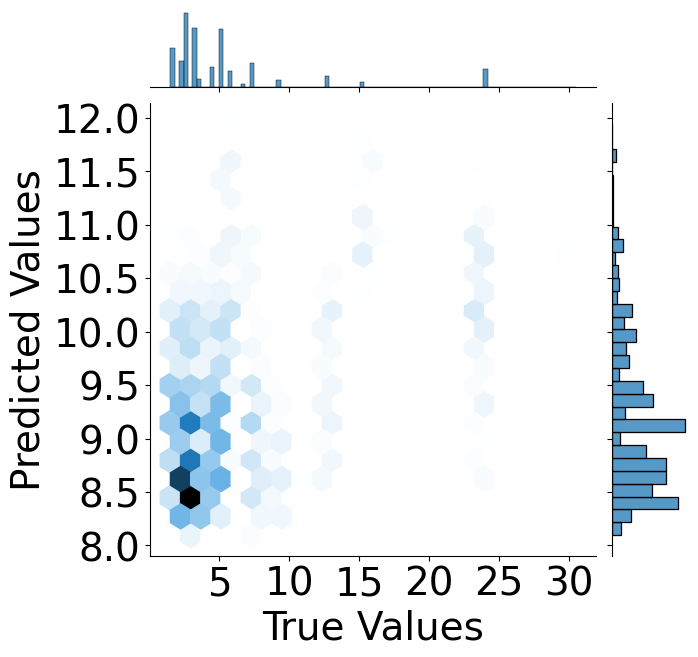}
     \\
     \makebox[0.19\textwidth][c]{(a) PollutionNet (NO$_2$)}
     \makebox[0.19\textwidth][c]{(b) CNN (NO$_2$)}
     \makebox[0.19\textwidth][c]{(c) LR (NO$_2$)} 
     \makebox[0.19\textwidth][c]{(d) XGBoost (NO$_2$)}
     \makebox[0.19\textwidth][c]{(e) LGBM (NO$_2$)}
     \\
     \includegraphics[width=0.19\textwidth]{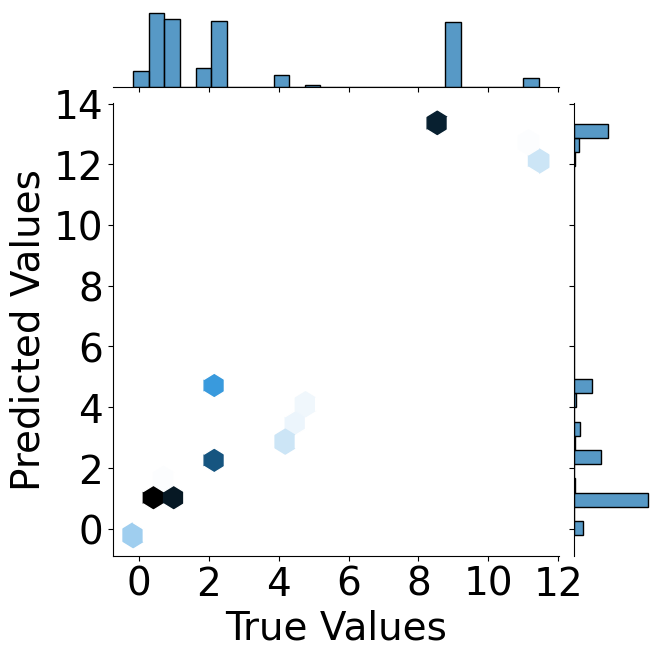}
     %\vspace{5pt}
     \includegraphics[width=0.19\textwidth]{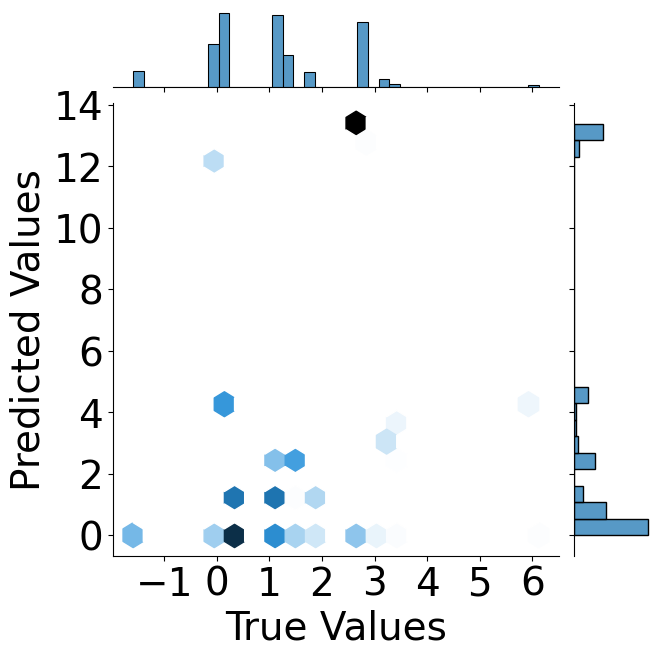}
     %\vspace{5pt}
     \includegraphics[width=0.19\textwidth]{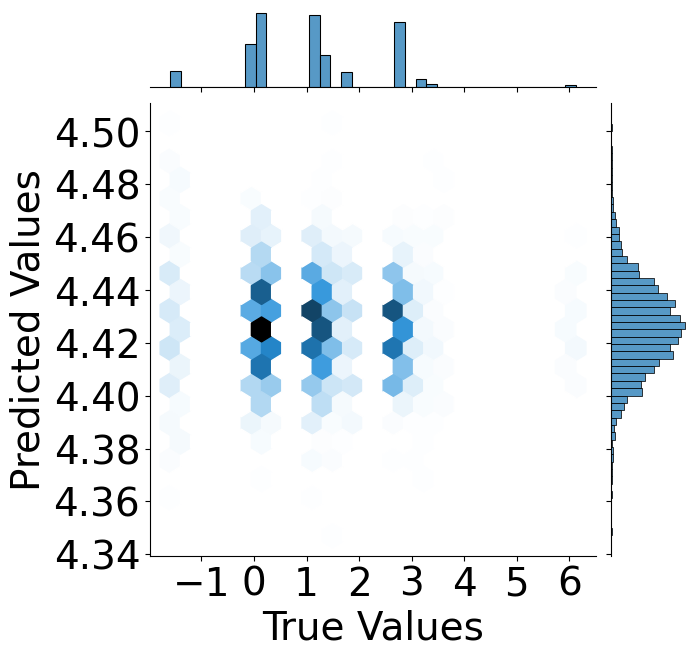}
     \includegraphics[width=0.19\textwidth]{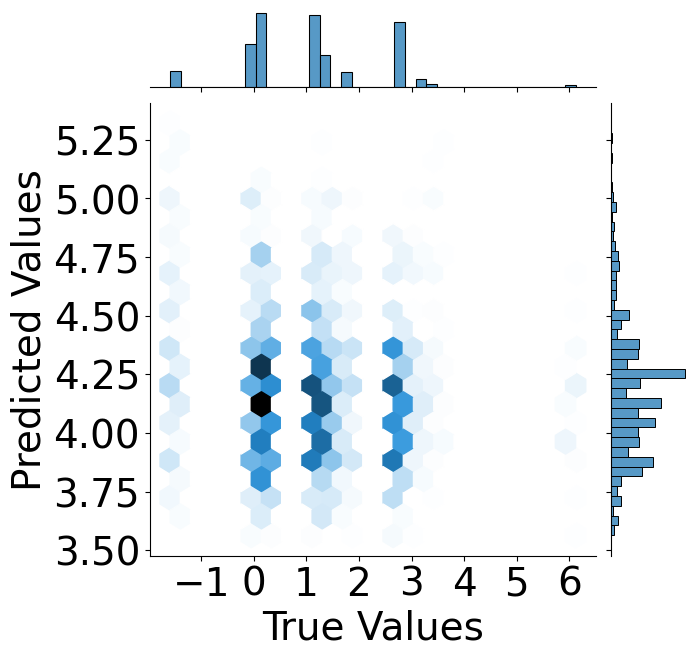}
     %\vspace{5pt}
     \includegraphics[width=0.19\textwidth]{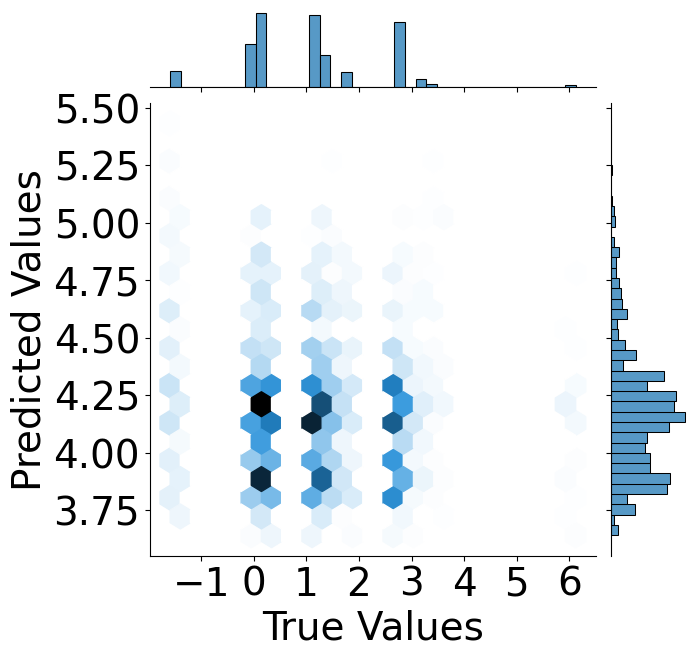}
     \\
     \makebox[0.19\textwidth][c]{(f) PollutionNet (NO$_2$)}
     \makebox[0.19\textwidth][c]{(g) CNN (NO$_2$)}
     \makebox[0.19\textwidth][c]{(h) LR (NO$_2$)} 
     \makebox[0.19\textwidth][c]{(i) XGBoost (NO$_2$)}
     \makebox[0.19\textwidth][c]{(j) LGBM (NO$_2$)} 

    \caption{Illustrating the intercomparison of the PollutionNet framework with CNN, linear regression, XGBoost, and LGBM models for NO$_2$ and SO$_2$ concentrations. The first row represents the comparison of NO$_2$, and the last row represents the comparison of SO$_2$. The color gradient, ranging from deep blue to light blue, represents the density of data points. Additionally, the outer line plot denotes the density area.}
    \label{fig:jointdist}
\end{figure}

A joint distribution analysis (Fig.~\ref{fig:jointdist}) further validates PollutionNet’s superiority. The scatter plots reveal that PollutionNet’s predictions (blue dots) align closely with the ideal regression line, indicating high agreement with ground-truth measurements. In contrast, CNN, LR, XGBoost, and LGBM exhibit greater dispersion, particularly at higher concentrations, where they tend to underpredict. The density plots (outer contours) confirm that PollutionNet’s errors are more tightly clustered around the true values, whereas other models show broader deviations.

\subsection{Robustness to Dataset Size Variations}

\begin{figure}[!ht]
    \centering
    \footnotesize
    \includegraphics[width=0.48\linewidth]{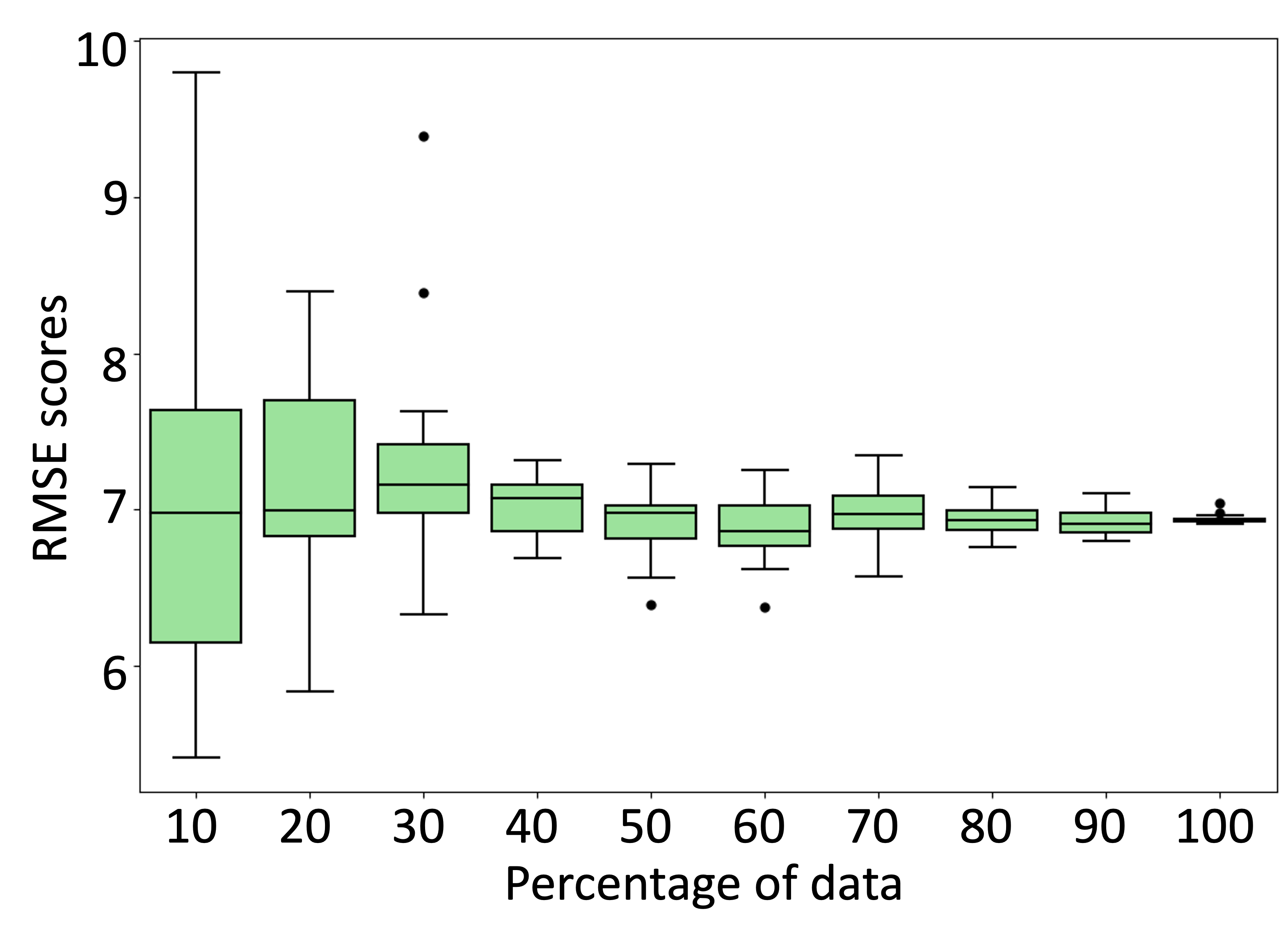}
     %\vspace{5pt}
     \includegraphics[width=0.48\linewidth]{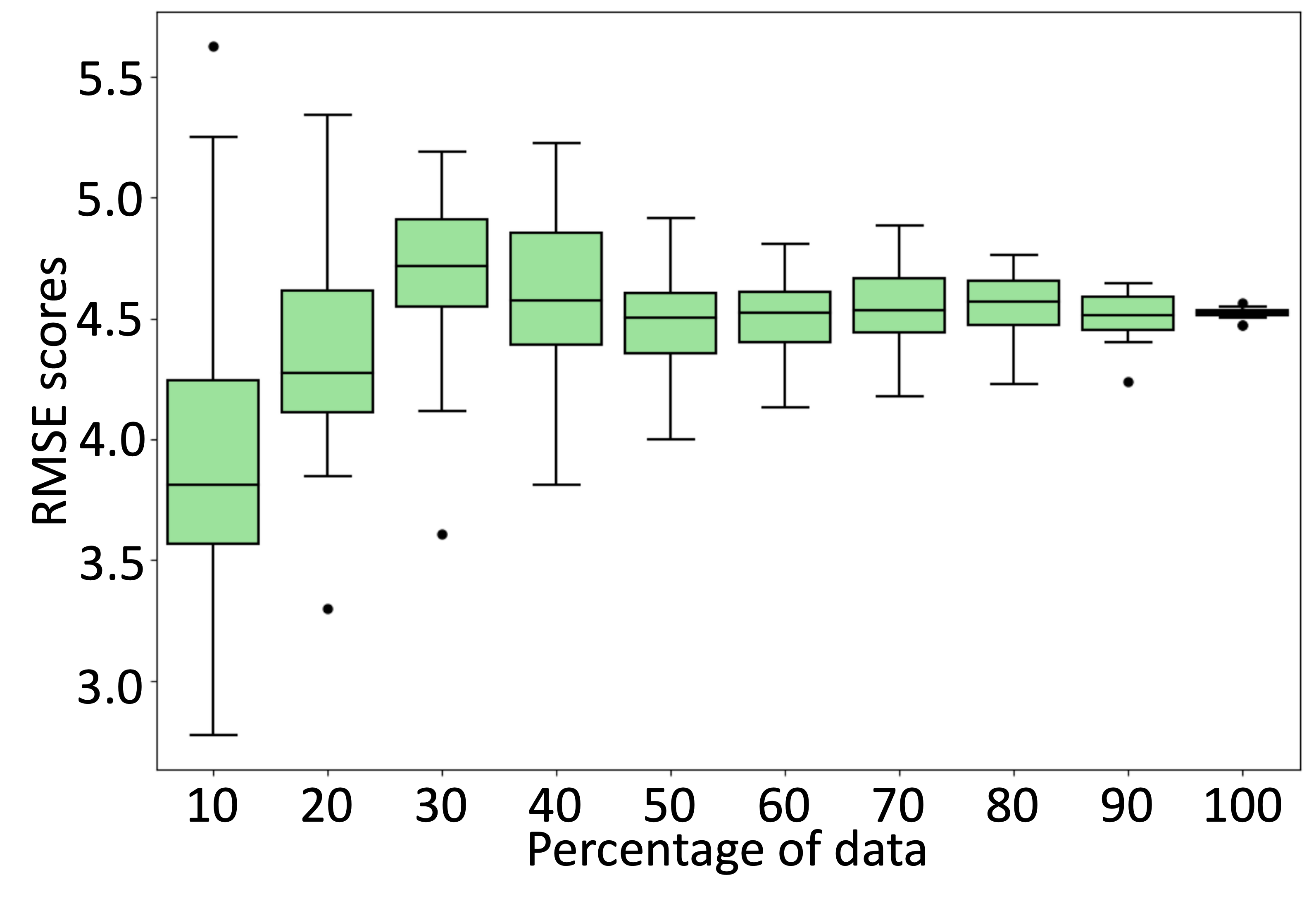}
     %\vspace{5pt}
     \\
     \makebox[0.46\linewidth][c]{(a) NO$_2$}
     \makebox[0.46\linewidth][c]{(b) SO$_2$}

    \caption{Visualization of RMSE scores for various percentages of dataset sizes for (a) NO$_2$, and (b) SO$_2$.}
    \label{fig:datasize}
\end{figure}

We evaluated PollutionNet’s sensitivity to training data volume by progressively increasing the dataset size from 10\% to 100\% (Fig.~\ref{fig:datasize}). The RMSE for both NO\textsubscript{2} and SO\textsubscript{2} remains stable across different data fractions, with only minor fluctuations (NO\textsubscript{2}: 6.92–7.26~\textmu g/m\textsuperscript{3}, SO\textsubscript{2}: 3.95–4.65~\textmu g/m\textsuperscript{3}). This suggests that PollutionNet performs well even with limited training samples, making it suitable for regions with sparse monitoring infrastructure.

\subsection{Benchmarking Against Recent Studies}

\renewcommand\theadfont{\normalsize\bfseries}
\renewcommand\theadalign{tl}

\begin{table}[!ht]
\scriptsize
\centering
\caption{Comparative analysis of PollutionNet with recent studies showing region, period, satellite data, resolution, model, and RMSE (in $\mu g/m^3$).}
\begin{tabular}{>{\raggedright\arraybackslash}p{2.2cm} >{\raggedright\arraybackslash}p{1.7cm} >{\raggedright\arraybackslash}p{1.6cm} >{\raggedright\arraybackslash}p{2.2cm} >{\centering\arraybackslash}p{1.4cm} >{\centering\arraybackslash}p{1.6cm} >{\raggedright\arraybackslash}p{2.3cm}}
\toprule
\textbf{Literature} & \textbf{Region} & \textbf{Period} & \textbf{Satellite Data} & \textbf{Res.} & \textbf{Model} & \textbf{RMSE ($\mu g/m^3$)} \\
\midrule
\cite{li2019satellite} & China & 2014--2015 & SO\textsubscript{2} from OMI & 0.25° & RF-STK & SO\textsubscript{2}: 10.36 \\
\cite{zhang2020estimating} & E. China & 2014 & NO\textsubscript{2}, CH\textsubscript{2}O from OMI & 0.25° & GWR & -- \\
\cite{wang2021estimating} & China & 2018--2020 & S5P-TROPOMI (O\textsubscript{3}, NO\textsubscript{2}) & 0.05°–0.07° & LightGBM & NO\textsubscript{2}: 8.44, O\textsubscript{3}: 17.7 \\
\cite{wei2023ground} & China & 2013--2020 & NO\textsubscript{2} from OMI & 0.25° & Decision Tree & NO\textsubscript{2}: 11.5 \\
\cite{xu2023downward} & BTH, China & 2014--2019 & NO\textsubscript{2} from OMI & 0.25° & -- & Avg NO\textsubscript{2}: 13.3 \\
\cite{zhu2023leso} & China, EU, USA & 2012--2021 & TROPOMI O\textsubscript{3} & 0.1° & Deep Forest & O\textsubscript{3}: 19.6 \\
\textbf{PollutionNet (This Study)} & Ireland & 2020--2021 & TROPOMI (NO\textsubscript{2}, SO\textsubscript{2}) & 0.05° & ViT & NO\textsubscript{2}: 6.89, SO\textsubscript{2}: 4.49 \\
\bottomrule
\end{tabular}
\label{tab:comparison}
\end{table}
  
A comparative review of recent air pollution prediction studies (Table~\ref{tab:comparison}) highlights PollutionNet’s advancements. In terms of spatial resolution, PollutionNet leverages TROPOMI satellite data at 0.05\textdegree\ resolution, surpassing previous studies that relied on coarser datasets ($\geq$ 0.07\textdegree). With respect to accuracy, our framework achieves lower RMSE values (6.89~\textmu g/m\textsuperscript{3} for NO\textsubscript{2}, 4.49~\textmu g/m\textsuperscript{3} for SO\textsubscript{2}) compared to state-of-the-art methods, such as RF-STK (10.36~\textmu g/m\textsuperscript{3} for SO\textsubscript{2}) and LightGBM (8.44~\textmu g/m\textsuperscript{3} for NO\textsubscript{2}). Furthermore, PollutionNet’s ViT backbone demonstrates superior performance over traditional architectures, including random forests, decision trees, and gradient boosting, thereby advancing the methodological frontier in atmospheric pollution prediction.

\subsection{Key Findings and Implications}  
PollutionNet outperforms conventional models in predicting NO\textsubscript{2} and SO\textsubscript{2} concentrations, achieving higher accuracy and improved spatial pattern recognition. The framework also exhibits temporal stability, reliably tracking pollutant trends over time. In addition, it is data-efficient, demonstrating robust performance even with limited training samples. When compared to recent studies, PollutionNet provides higher-resolution predictions and lower error rates, thereby establishing itself as a viable tool for environmental monitoring. These findings highlight PollutionNet’s potential for real-world deployment in regions lacking dense air quality monitoring networks, offering policymakers a reliable tool for pollution assessment and mitigation.

%-------------------------------------------------------------------------------------------

\section{Conclusion and Future Scope}
\label{conclusion}
The study presents PollutionNet, a ViT-based framework for predicting near-surface NO\textsubscript{2} and SO\textsubscript{2} concentrations by integrating satellite and ground-based data. Leveraging ViT's self-attention mechanism, the model effectively captures spatiotemporal dependencies, outperforming traditional approaches with RMSE scores of 6.89 \textmu g/m$^3$ (NO\textsubscript{2}) and 4.49 \textmu g/m$^3$ (SO\textsubscript{2}). This advancement addresses critical gaps in air quality monitoring, offering a reliable tool for policymakers to mitigate pollution impacts. The fusion of Sentinel-5P TROPOMI satellite data with ground observations ensures robust predictions, even in data-scarce regions, thereby enhancing public health strategies and environmental management.

Future research directions include: (1) expanding PollutionNet's geographical coverage by adapting it to diverse regions with localized datasets; (2) incorporating additional pollutants like PM\textsubscript{2.5} and O\textsubscript{3} for comprehensive air quality assessment; (3) developing advanced imputation techniques to handle missing satellite data more effectively; (4) utilizing higher temporal resolution inputs to improve short-term forecasting accuracy. These enhancements would significantly strengthen PollutionNet's capability to support global environmental sustainability initiatives.
%------------------------------------------------------------------------------------------

\section*{Acknowledgment}
This research was funded by the Research Ireland Centre for Research Training in Digitally-Enhanced Reality (d-real) under Grant No. 18/CRT/6224. This research was conducted with the financial support of Research Ireland Centre under Grant Agreement No.\ 13/RC/2106\_P2 at the ADAPT Research Ireland Centre at University College Dublin. ADAPT, the Research Ireland Centre for AI-Driven Digital Content Technology, is funded by Research Ireland Centre.

\section*{Author Contributions}

P.D. developed the concept, implemented the methodology, and wrote the main manuscript text. S.D. contributed to the design of the deep learning model and supervised the experimental setup. B.S.P. provided critical revisions, manuscript structuring guidance, and technical oversight. All authors reviewed the manuscript and approved the final version.

\section*{Declarations}

\textbf{Ethical responsibilities of authors:} \\
All authors have read, understood, and have complied as applicable with the statement on “Ethical responsibilities of Authors” as found in the Instructions for Authors.

\noindent \textbf{Competing Interests:} \\
The authors declare that they have no competing interests.

\noindent \textbf{Funding:} \\
This research was funded by the Research Ireland Centre for Research Training in Digitally-Enhanced Reality (d-real) under Grant No. 18/CRT/6224, and by the ADAPT Research Centre at University College Dublin under Grant Agreement No. 13/RC/2106 P2.

\noindent \textbf{Data Availability:} \\
The data that support the findings of this study are available from the corresponding author upon reasonable request.

\noindent \textbf{Ethical Approval:} \\
Not applicable.

\noindent \textbf{Consent to Participate:} \\
Not applicable.

\noindent \textbf{Consent to Publish:} \\
Not applicable.

\bibliography{sn-bibliography}% common bib file

\end{document}